\documentclass{article}

\usepackage[preprint]{neurips_2025}

\usepackage[utf8]{inputenc} 
\usepackage[T1]{fontenc}    
\usepackage{hyperref}       
\usepackage{url}            
\usepackage{booktabs}       
\usepackage{amsfonts}       
\usepackage{nicefrac}       
\usepackage{microtype}      
\usepackage{graphicx}
\usepackage{multirow}
\usepackage{subfloat}
\usepackage[caption=false,font=footnotesize,labelfont=rm,textfont=rm]{subfig}

\usepackage[table]{xcolor}
\definecolor{lightred}{rgb}{1, 0.85, 0.85}  
\definecolor{lightblue}{rgb}{0.85,0.95,1} 
\definecolor{first}{HTML}{547CB1} %
\definecolor{improve}{HTML}{1E73C4} %

\title{\raisebox{-0.085cm}{\includegraphics[scale=0.03]{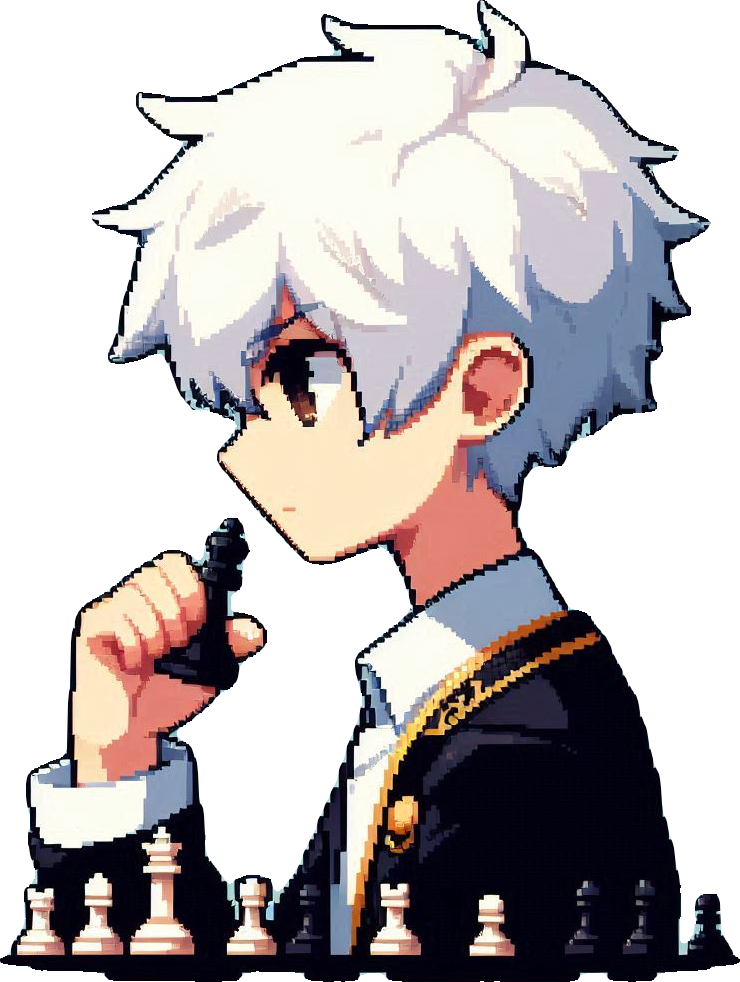}}~Who is a Better Player: LLM against LLM}

\author{Yingjie Zhou~~~Jiezhang Cao~~~Farong Wen~~~Li Xu~~~Yanwei Jiang~~~Jun Jia~~~Ronghui Li \\\textbf{Xiaohong Liu ~~~Yu Zhou ~~~Xiongkuo Min ~~~Jie Guo~~~Zicheng Zhang~~~Guangtao  Zhai} \\
\textsuperscript{\rm 1} Shanghai Jiao Tong University 
\hspace{0.3cm} \textsuperscript{\rm 2} PengCheng Laboratory \hspace{0.3cm} \textsuperscript{\rm 3} Harvard Medical School
 \\ 
\textsuperscript{\rm 4} Shanghai AI Laboratory \hspace{0.3cm} \textsuperscript{\rm 5} China University of Mining and Technology
}

\begin{document}

\maketitle

\begin{figure*}[!h]
    \vspace{-1.1cm}
    \centering
    \includegraphics[width =0.98\linewidth]{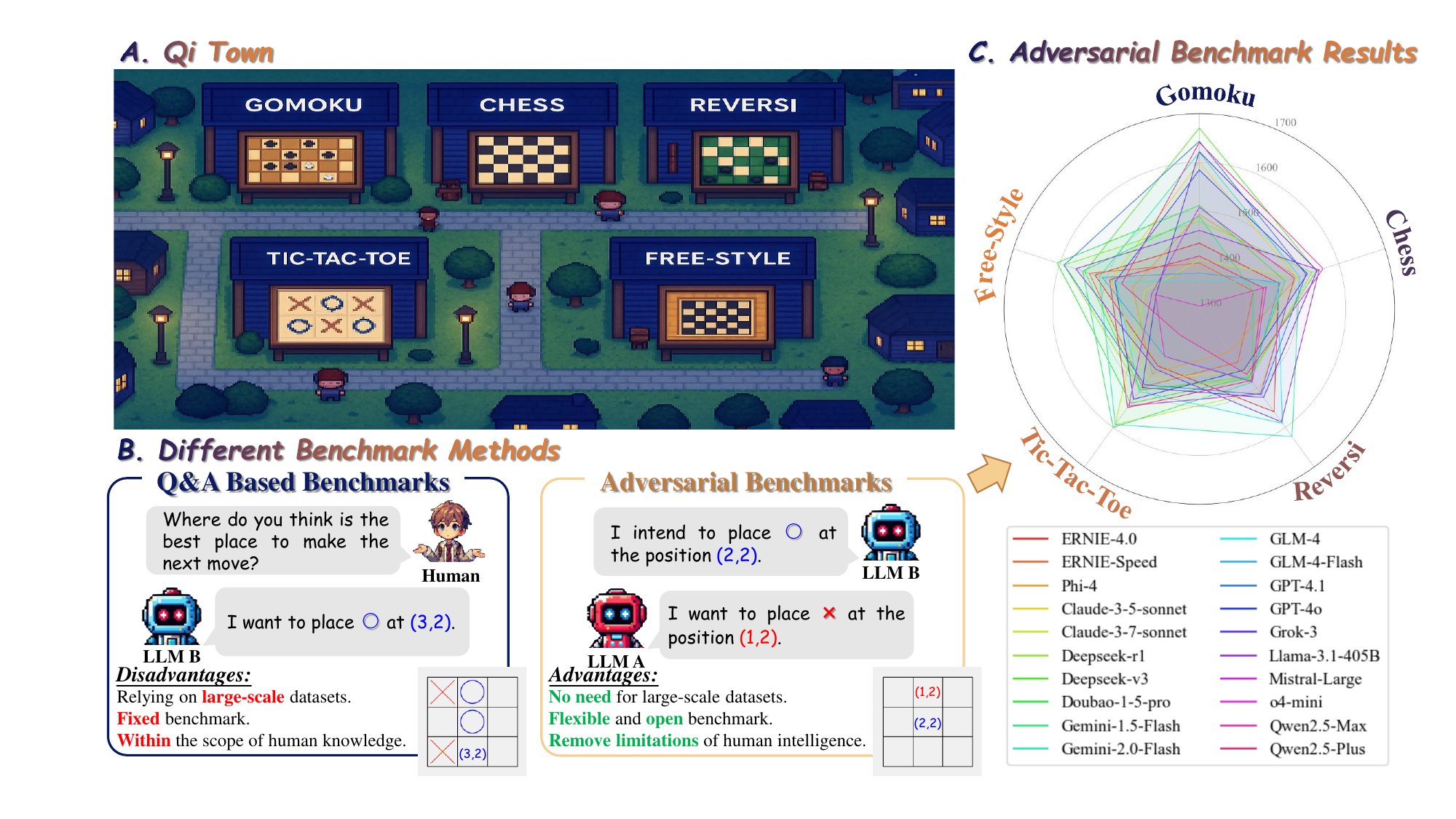}
    \vspace{-0.4cm}
    \caption{The implementation, advantages and results of our adversarial benchmark. Subfigure A shows the Qi Town proposed in this paper, which provides a basic platform for adversarial benchmarks; Subfigure B compares the differences between Q\&A-based benchmark schemes and our adversarial benchmark; Subfigure C shows the performance measurement results of different LLMs.}
    \label{fig:teaser}
\end{figure*}

\begin{abstract}

Adversarial board games, as a paradigmatic domain of strategic reasoning and intelligence, have long served as both a popular competitive activity and a benchmark for evaluating artificial intelligence (AI) systems. Building on this foundation, we propose an adversarial benchmarking framework to assess the comprehensive performance of Large Language Models (LLMs) through board games competition, compensating the limitation of data dependency of the mainstream Question-and-Answer (Q\&A) based benchmark method. We introduce \textbf{Qi Town}, a specialized evaluation platform that supports 5 widely played games and involves 20 LLM-driven players. The platform employs both the Elo rating system and a novel Performance Loop Graph (PLG) to quantitatively evaluate the technical capabilities of LLMs, while also capturing Positive Sentiment Score (PSS) throughout gameplay to assess mental fitness. The evaluation is structured as a round-robin tournament, enabling systematic comparison across players. Experimental results indicate that, despite technical differences, most LLMs remain optimistic about winning and losing, demonstrating greater adaptability to high-stress adversarial environments than humans. On the other hand, the complex relationship between cyclic wins and losses in PLGs exposes the instability of LLMs' skill play during games, warranting further explanation and exploration.


\end{abstract}

\section{Introduction}
Board games have long been regarded as intellectually demanding competitions that assess a broad spectrum of cognitive and affective skills, including memory, logical reasoning, calculation, strategic planning, risk anticipation, and emotional regulation. Their enduring popularity worldwide stems not only from their competitive nature but also from their capacity to holistically evaluate human intelligence. With the rapid advancement of artificial intelligence (AI), these traditionally human-versus-human games have increasingly become arenas for human-computer interaction.

Recently, the emergence of large language models (LLMs) has positioned them as novel participants in such games, opening up new opportunities for comprehensive LLM evaluation. Unlike conventional assessment methods that typically isolate specific cognitive capabilities, strategic board games offer a richer, multifaceted environment to evaluate LLMs’ integrated competencies. These include game memory, move reasoning and analysis, opponent behavior prediction, and emotional stability in dynamic, adversarial settings. To harness these properties, we propose an adversarial evaluation paradigm—LLM against LLM—as a new framework for LLM benchmarking. As shown in Fig.~\ref{fig:teaser}, in contrast to standard question-and-answer (Q\&A)-based benchmarks that rely on static, human-curated corpora, adversarial benchmark enables continuous, self-generated assessment content through real-time interactions between models. Overall, the proposed paradigm offers several notable advantages: 1) \textbf{Dynamic benchmarking} is achieved through diverse and evolving game states; 2) \textbf{Environment awareness} is considered through the inclusion of high-stress adversarial conditions that challenge LLM adaptability; 3) \textbf{Safety and reliability} are enhanced by a dynamic evaluation mechanism.

In this work, we construct Qi Town as shown in Fig.~\ref{fig:teaser}, undertaking a comprehensive investigation of LLM performance in adversarial board game settings. Specifically, we design experiments across five games: Gomoku, Chess, Reversi, Tic-Tac-Toe—representing traditional rule-based games—and a novel Free-Style game mode. In the Free-Style setup, LLMs collaboratively determine the rules before initiating play, thus testing their autonomy and negotiation capabilities. Across all games, we employ the Elo rating system to assess competitive skill, and we introduce a Performance Loop Graph (PLG) to visualize LLM behaviors across multiple matches. To evaluate emotional aspects, we record sentiment changes during gameplay and compute a Positive Sentiment Score (PSS) to capture LLMs’ affective responses. The principal contributions of this work are summarized as follows:

\begin{itemize}
\item \textbf{Adversarial Benchmarking Platform:} We present Qi Town, a novel evaluation framework supporting 20 LLMs engaged in board game competition. This platform introduces an adversarial benchmarking paradigm that surpasses the limitations of static Q\&A-based evaluations and reduces reliance on human-generated data.
\item \textbf{Diverse Game Set and Multidimensional Metrics:} The benchmark includes both fixed-rule games and a Free-Style mode that allows models to co-construct game rules. We evaluate performance across both technical (Elo ratings, PLG) and psychological (PSS) dimensions, providing a multidimensional view of LLM capabilities.
\item \textbf{Comprehensive Experimental Analysis:} The experimental results reveal limitations in some LLMs’ reasoning, decision-making capabilities, and possible risk of affective tendencies toward negative emotion. However, most LLMs can adapt to high-stress competition with optimism. Besides, the cyclic win–loss patterns observed in PLGs offer novel insights into the structural stability and relative performance dynamics of LLMs in competitions.
\end{itemize}

\section{Related Work}
\label{rw}

\subsection{Q\&A-based Benchmarks for LLM}
Benchmarks play a critical role in both validating and advancing the capabilities of LLMs. Currently, the predominant evaluation paradigm is Q\&A-based, which involves constructing large-scale evaluation corpora and assessing LLMs' performance based on their responses. Typically, benchmarks like C-Eval \cite{huang2023c}, MMLU \cite{mmlu}, and Big-Bench \cite{bigbench} are designed to assess knowledge comprehension and memorization, while others such as TNEWS \cite{tnews}, IFLYTEK \cite{tnews}, ChID \cite{tnews}, and CMRC2018 \cite{cui-etal-2019-span} focus on evaluating natural language processing skills. This Q\&A-based approach has also been extended to multimodal LLMs (MLLMs), where benchmarks such as MME and MMBench assess perceptual and reasoning abilities, MER-Bench \cite{lian2024merbench} and MEMO-Bench \cite{zhou2024memo} emphasize affective comprehension, Q-Bench \cite{wu2023q}, Q-Bench-Video \cite{zhang2024q}, and A-Bench \cite{zhang2024bench} evaluate visual quality understanding, and VSI-Bench \cite{yang2024thinking} targets visuospatial perception.

Despite the diversity and depth of these benchmarks, Q\&A-based evaluation schemes face several inherent limitations. Constructing large-scale, high-quality benchmarks is resource-intensive, requiring substantial time, cost, and manual annotation. Furthermore, because ground truth labels are constrained by human cognition, the scope of such evaluations is fundamentally limited to human-level intelligence. As LLMs continue to evolve and begin to exhibit behaviors and competencies that potentially exceed those of human evaluators, these constraints are becoming increasingly apparent. This highlights the urgent need to explore novel benchmark paradigms that move beyond traditional Q\&A frameworks to more fully capture the emerging capabilities of advanced LLMs.

\subsection{AI in Games}
Over the past decade, the advent of reinforcement learning (RL) has catalyzed significant breakthroughs in the application of AI to gaming. In competitive game environments, RL-based agents such as AlphaStar \cite{vinyals2019grandmaster} and OpenAI Five \cite{berner2019dota} have demonstrated superhuman performance in complex real-time strategy games like StarCraft II and Dota 2, respectively, by leveraging layered model architectures and self-play training paradigms. In the domain of board games, AlphaGo \cite{silver2016mastering} marked a transformative milestone by integrating Monte Carlo Tree Search (MCTS) with dual deep neural networks to defeat the world champion in Go in 2016. Its successor, AlphaGo Zero \cite{silver2017mastering}, further advanced this approach by relying entirely on self-play without human supervision.

In recent years, LLMs have emerged as promising agents in various game-related tasks. For example, AI Town, proposed by Park $et$ $al.$ \cite{park2023generative}, introduced a virtual community of 25 LLM-based agents, showcasing initial explorations of LLMs in interactive, multi-agent environments. Wang $et$ $al.$ \cite{wang2025large} examined the gaming capabilities of MLLMs using puzzles such as Sudoku and Minesweeper, as well as tasks involving search algorithms, though their work did not fully exploit the competitive and strategic nature of multi-agent gameplay. Huang $et$ $al.$ \cite{huang2025competing} investigated the decision-making capabilities of LLMs through game-theoretic scenarios, such as the El Farol bar problem \cite{arthur1994inductive}, successfully integrating game theory into LLM evaluation. However, their analysis was limited to static, single-play games. To address these limitations, we propose a novel evaluation paradigm based on dynamic, strategy-rich board games. Specifically, we introduce a competitive setting wherein LLMs directly confront one another in a variety of board games characterized by explicit rules and complex decision spaces. This LLM against LLM framework is designed to comprehensively assess both the technical and psychological capabilities of state-of-the-art LLMs in adversarial scenarios.

\section{Qi Town}
\begin{figure*}[!h]
    \centering
    \includegraphics[width =\linewidth]{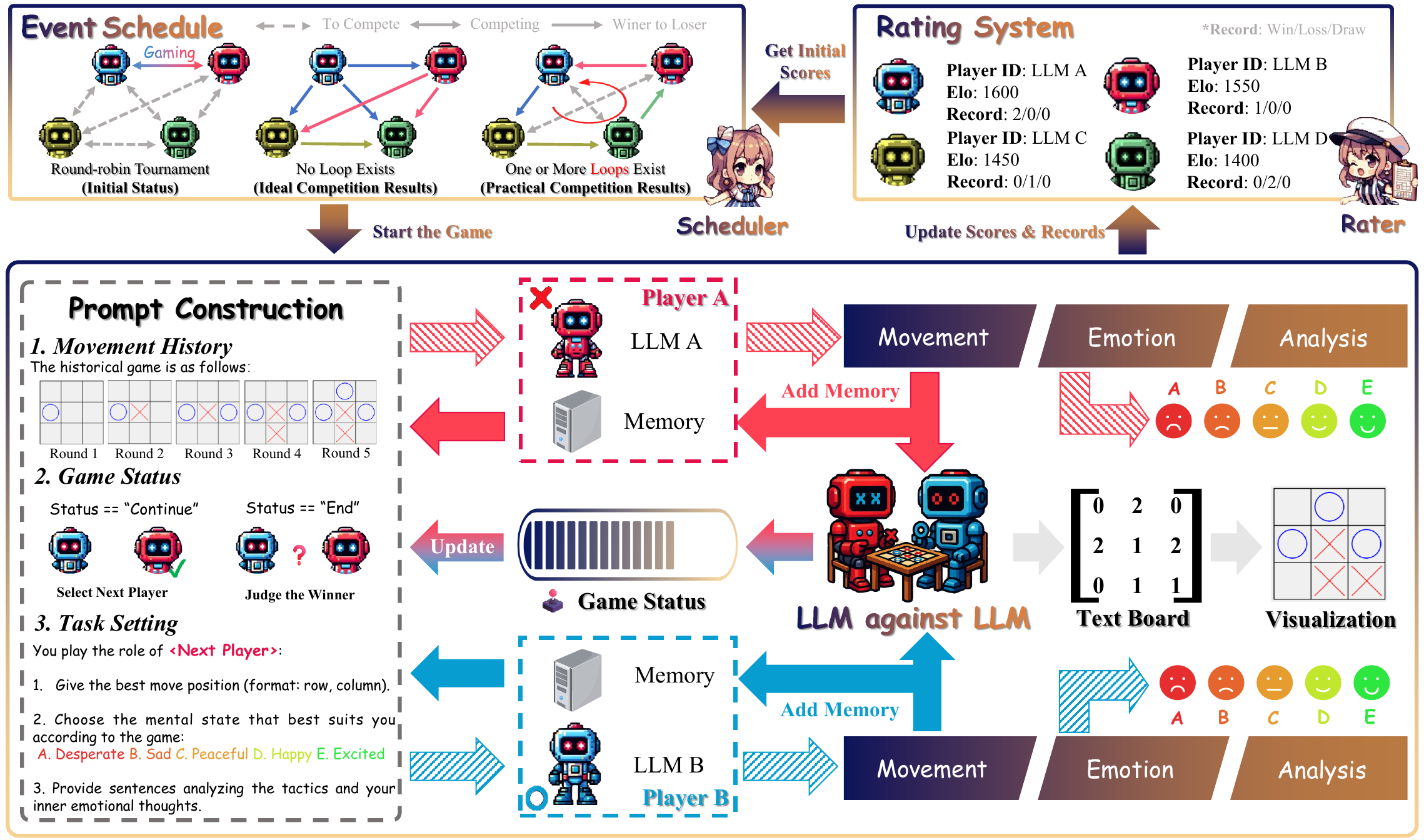}
    \vspace{-0.1cm}
    \caption{The framework of Qi Town, which consists of three parts: tournament scheduling, LLM against LLM and rating system.}
    \label{fig:framework}
    \vspace{-0.6cm}
\end{figure*}

\subsection{Overview}
Qi Town, as illustrated in Fig.~\ref{fig:framework}, is a virtual community specifically designed to facilitate competitive board games among LLMs. From a scheduling perspective, Qi Town is equipped to manage round-robin tournaments, enabling flexible organization of adversarial matchups. The core functionality of Qi Town lies in its game process control module, which orchestrates matches between LLMs in accordance with predefined game rules. During gameplay, the system continuously updates the game status and alternately prompts each LLM with the current game context. Based on these prompts, each player generates a response comprising three elements: the movement, emotion and a brief analysis. The move updates the game status and is logged, while the emotional and explanatory components are retained for post-game analysis. Besides, Qi Town provides automated visualization and storage of game sequences. This process iterates until a win or draw is reached, after which the system proceeds to the next scheduled match. For evaluation, Qi Town continuously tracks and updates each LLM's Elo score and associated performance metrics, ensuring a dynamic and quantitative assessment of both technical proficiency and emotional changes across games.

\subsection{Game Types}
\textbf{Fixed-Rule Games} are a class of deterministic, turn-based environments governed by fixed rules and clear win conditions, offering ideal testbeds for evaluating strategic reasoning and decision-making. This study considers four representative fixed-rule games of increasing complexity. Among all, Tic-Tac-Toe is a simple two-player game played on a 3×3 grid, where players alternate marking empty cells with “X” or “O” to align three symbols in a row. Gomoku expands this structure to a 15×15 board, requiring players to align exactly five consecutive stones of the same color. Reversi, conducted on an 8×8 grid, involves flipping the opponent’s discs by flanking them, aiming to control the majority of the board by the end of the game. Chess is a globally standardized strategy game played on an 8×8 board with 16 uniquely functioning pieces per player, requiring both deep tactical calculation and long-term planning to achieve checkmate under complex movement and game-state rules. These games span a spectrum of spatial, combinatorial, and cognitive demands, making them effective platforms for analyzing algorithmic strategies and LLMs' behavior in rule-bound settings.

\textbf{Free-Style} is an open-ended board game paradigm designed to assess negotiation and rule-making abilities in LLM against LLM settings. Played on a 5×5 board, the game does not begin with predefined rules. Instead, the two participating LLMs engage in an iterative negotiation process to co-construct a mutually agreed-upon rule set. Once finalized, gameplay proceeds according to these self-defined rules until a winner is determined. Free-Style removes human-imposed constraints from the evaluation process, offering a unique opportunity to study the rule formation, adaptation, and interactive negotiation capabilities of LLMs in a controlled adversarial context.

\subsection{Players: Large Language Models}
Distinct from prior studies on competitive board games, which typically involve human players or search-based algorithms as opponents, this work introduces a novel evaluation paradigm by setting both competing entities as LLMs. To ensure the diversity and representativeness of the participants, we included LLMs developed by twelve different organizations, selecting the recently released versions from each to maintain experimental timeliness. In total, 20 LLMs are chosen for participation in the chess tournament, including: Claude-3-7-sonnet \cite{anthropic2025claude}, Claude-3-5-sonnet \cite{anthropic2024claude}, Deepseek-v3 \cite{liu2024deepseek}, Deepseek-r1 \cite{guo2025deepseek}, Doubao-1.5-pro \cite{doubao}, ERNIE-Speed \cite{erniespeed}, ERNIE-4.0 \cite{ernie4}, Llama-3.1-405 \cite{llama}, Mistral-Large \cite{mistral}, Phi-4 \cite{abdin2024phi}, Grok-3 \cite{grok}, Gemini-1.5-Flash \cite{team2024gemini}, Gemini-2.0-Flash \cite{gemini2}, Qwen2.5-Plus \cite{yang2024qwen2}, Qwen2.5-Max \cite{yang2024qwen2}, GPT-4.1 \cite{gpt4}, GPT-4o \cite{gpt4o}, and o4-mini \cite{o4}. All LLMs are accessed via their official Application Programming Interfaces (APIs) to ensure consistency in interaction and scheduling. As illustrated in Fig.~\ref{fig:framework}, the gameplay interface is purely textual, with no visual processing involved; instead, board status is represented and interpreted using a standardized algebraic notation system, through which all move decisions are made.
\begin{figure*}[!t]
    \centering
    \includegraphics[width =\linewidth]{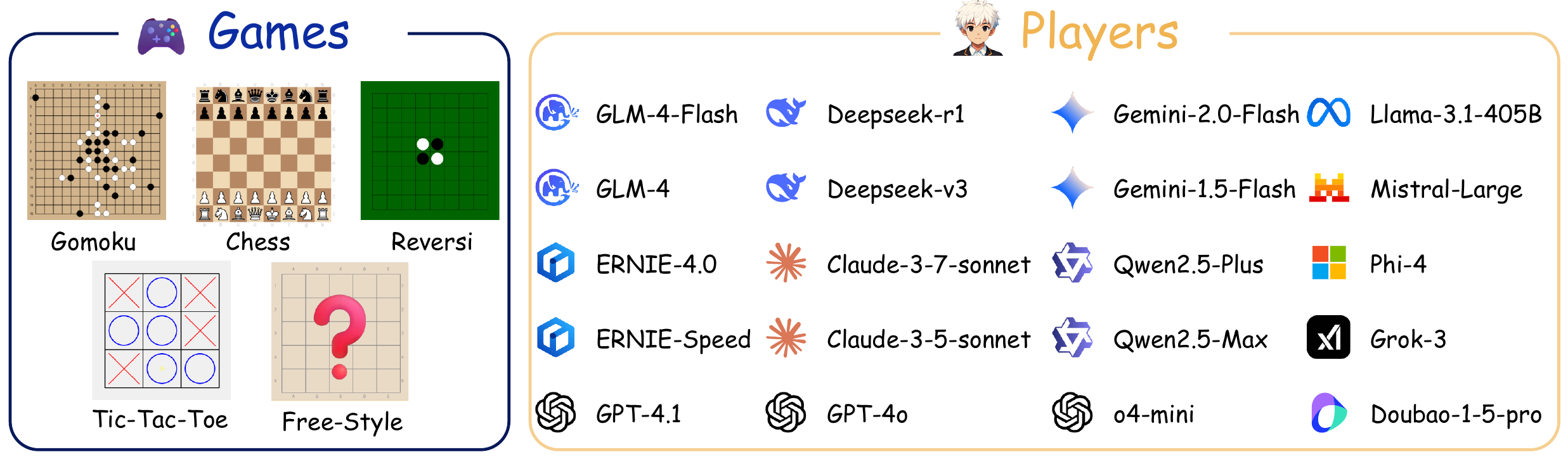}
    
    \caption{Visualization of various board games and a list of players.}
    \label{fig:gameplayer}
    \vspace{-0.5cm}
\end{figure*}
\subsection{Game Visualization}
While coordinate and algebraic notations can fully represent the progression of a game, they are not well-suited for further analysis due to their lack of visual intuitiveness. To facilitate a clearer understanding of the moves made by both sides' LLMs, we implement a visual representation of each move using Pygame \cite{shinde2021pygame} and Python-chess \cite{fiekas2024pythonchess}, as illustrated in Fig.~\ref{fig:gameplayer}. Additional visualizations, including examples from newly introduced Free-Style game types, are presented in Sec.~\ref{sec:games}.
\subsection{Elo Rating System}
The Elo rating system \cite{elo2008rating}, is a widely adopted statistical method for evaluating player performance across various competitive domains. Originally designed for chess, it has since been applied to other disciplines such as Chinese Chess, Go, football, basketball, and eSports, and is now regarded as a standard for quantifying relative skill levels. In this study, we employ the Elo scoring system to evaluate the overall performance of 20 players across a series of chess matchmaking tournaments. The core assumption of the Elo model is that a player’s skill level follows a normal distribution centered around their mean performance. The probability density functions for two competing players $X$ and $Y$ with average performance scores $U_x$ and $U_y$ can be described as:
\begin{equation}
\begin{array}{l}
f(x) = \frac{1}{{\sqrt {2\pi } \delta }}\exp \left(\frac{{ - {{(x - {U_x})}^2}}}{{2{\delta ^2}}}\right),\\
f(y) = \frac{1}{{\sqrt {2\pi } \delta }}\exp \left(\frac{{ - {{(y - {U_y})}^2}}}{{2{\delta ^2}}}\right),
\end{array}
\end{equation}
where $\delta$ denotes the standard deviation, reflecting the same stability of each player's performance. Based on this assumption, the expected probability of player $X$ defeating player $Y$, given their performance scores are random variables $x$ and $y$ respectively, is computed as:
\begin{equation}
\label{equ2}
P(D) = P(x - y) = \frac{1}{2} + \int_0^D {\frac{1}{{\sqrt {2\pi } \sigma }}\exp \left(\frac{{ - {x^2}}}{{2{\sigma ^2}}}\right)} dx, \sigma  = \sqrt 2 \delta,
\end{equation}
where $D$ denotes the difference in performance scores between the two competitors. Since the integral form in Eq.~(\ref{equ2}) is computationally intensive, an approximation is commonly used in practice, derived through curve fitting using the least squares method. This yields the following sigmoid-like function:
\begin{equation}
\label{equ3}
    P(D) = \frac{1}{{1 + {{10}^{ - \frac{D}{{400}}}}}}.
\end{equation}
It is worth stating that the constants in Eq.~(\ref{equ3}) are empirically derived to closely match the theoretical distribution in Eq.~(\ref{equ2}), providing a computationally efficient alternative for estimating the expected win probability $P(D)$. Finally, given the Elo scores of players $X$ and $Y$ before $n$-th ($n \in {N^ + }$) match, denoted as $E{S_X}(n - 1)$ and $E{S_Y}(n - 1)$, their updated scores after the match are calculated as:
\begin{equation}
    \begin{array}{l}
E{S_X}(n) = E{S_X}(n - 1) + K({W_X} - P(D)),\\
E{S_Y}(n) = E{S_Y}(n - 1) + K({W_Y} - P(D)),
\end{array}
\end{equation}
where $K$ is a scaling factor that controls the sensitivity of the rating updates, and $W_X$, $W_Y$ represent the actual match outcomes. Specifically, a win yields $W=1$, a loss $W=0$ and a draw $W=0.5$ for both players. This update process is repeated for each match in the tournament series, ultimately yielding the final Elo scores for all 20 players.

\subsection{Performance Loop Graph}
While the Elo rating system provides an aggregate measure of each player's overall performance, it does not capture the relational dynamics between individual competitors. To address this limitation, we introduce a Performance Loop Graph (PLG), a directed graph-based visualization that encodes the win–loss relationships among players. Formally, the PLG is defined as:
\begin{equation}
    G = \{ (V,E)|v \in V,e \in E\},\\
\end{equation}
where each player is represented as a vertex $v$, and every directed edge $e$ represents a match outcome. Specifically, an edge is directed from the winner to the loser, indicating the result of a head-to-head match. In cases of a draw, no edge is added between the corresponding nodes. The set 
$V$ thus comprises all participating players, and $E$ contains all directed edges formed based on match outcomes. This graph structure allows for the application of graph-theoretic metrics to gain deeper insights into player performance. For each vertex $v$, the out-degree ${d^ + }(v)$ corresponds to the number of wins, while the in-degree ${d^ - }(v)$ denotes the number of losses. These metrics provide a local view of each player's performance in relation to others. Moreover, the presence of loops within the directed graph $G$ signifies mutual victories among groups of players, reflecting competitive balance and performance volatility. Of particular interest is the largest loop in $G$, which includes the maximum number of players. This loop is indicative of the most competitive and unstable subset of the player pool, offering valuable insight into the intensity and unpredictability of match outcomes.

\subsection{Positive Sentiment Score}
In contrast to emotions elicited through traditional human-centered methods, the spontaneous emotional responses exhibited by LLMs during task execution may offer deeper insights into their intrinsic behavioral traits, particularly their capacity for resilience under adversity. This characteristic is crucial for ensuring that LLMs maintain a stable and constructive affective state, thereby supporting more harmonious human–machine interactions.

To capture these emotional dynamics, we implemented a self-reporting protocol wherein players are asked to select their emotional state after each round of gameplay from the following options: \textit{A. Desperate, B. Sad, C. Peaceful, D. Happy, and E. Excited}. To quantitatively assess the overall emotional positivity of the players, we introduce the Positive Sentiment Score (PSS), a metric based on the expected value of a discrete random variable $Z$, which encodes emotional states. The PSS is computed as follows:

\begin{equation}
E(Z) = \sum\limits_{i \in I} {{p_i}{z_i}} ,I = \{ A,B,C,D,E\} , 
\end{equation}
where $I$ denotes the set of possible emotional categories, ${z_i} \in \{  - 2, - 1,0,1,2\} $ represents the sentiment score assigned to each emotion, and $p_i$ denotes the empirical probability of a player reporting emotion $i$. A higher PSS indicates a greater tendency toward positive emotional expression, thereby providing a useful quantitative measure of the emotional stability and optimism exhibited by the LLMs during gameplay.

\subsection{Competition Schedule: Experimental Setup}
To comprehensively evaluate the overall capabilities of individual players, a round-robin tournament structure is employed across five distinct game types. Specifically, for each game type, every player competes against each of the other 19 players exactly once, resulting in a total of 190 matches per full round-robin cycle. To mitigate potential biases introduced by variations in player performance or match scheduling order, each tournament is repeated three times with independently randomized match sequences. The final Elo scores, performance metrics, and PSSs for each player are computed by averaging the results across the three independent tournaments. It is noteworthy that the order of matches in each round-robin cycle is fully randomized to ensure fairness and minimize ordering effects. In total, the evaluation encompasses 2,850 = 3 × 190 × 5 games.

\begin{minipage}[!t]{0.49\textwidth}
\makeatletter\def\@captype{table}
\renewcommand{\arraystretch}{0.7}
\resizebox{0.95\linewidth}{!} {\begin{tabular}{ccc}
    \toprule
    Games     & Number of Rounds     & Rounds Range \\
    \midrule
    Gomoku &   5,222 & [9, 129]   \\
    Chess     &  28,931 & [11,299]     \\
    Reversi     &  11,397 & \{60\}\\
    Tic-Tac-Toe     & 1,343       & [5, 9]  \\
    Free-Style     & 1,904 & [5, 18]  \\
    \bottomrule
    \vspace{-0.3cm}
\end{tabular}}
\caption{Average total number of rounds and range of rounds for different classes of games. Values have been rounded.}
\label{tab:rounds}
\end{minipage}
~~
\begin{minipage}[!t]{0.49\textwidth}
\makeatletter\def\@captype{table}
\renewcommand{\arraystretch}{0.7}
\resizebox{0.95\linewidth}{!} {\begin{tabular}{ccc}
    \toprule
    Games     & Number of Loops     & Maximum Loop \\
    \midrule
    Gomoku &  93 & 17     \\
    Chess     & 8 & 6       \\
    Reversi     & 79 & 17  \\
    Tic-Tac-Toe     & 79 & 19  \\
    Free-Style     & 93 & 19 \\
    \bottomrule
    \vspace{-0.3cm}
\end{tabular}}
\caption{The average number of loops and nodes covered by maximum loop included in PLGs for different games. Values have been rounded.}
\label{tab:loops}
\end{minipage}

\begin{figure}[htbp]
	\centering
	\begin{minipage}{0.48\linewidth}
		\centering
		\includegraphics[width=\linewidth]{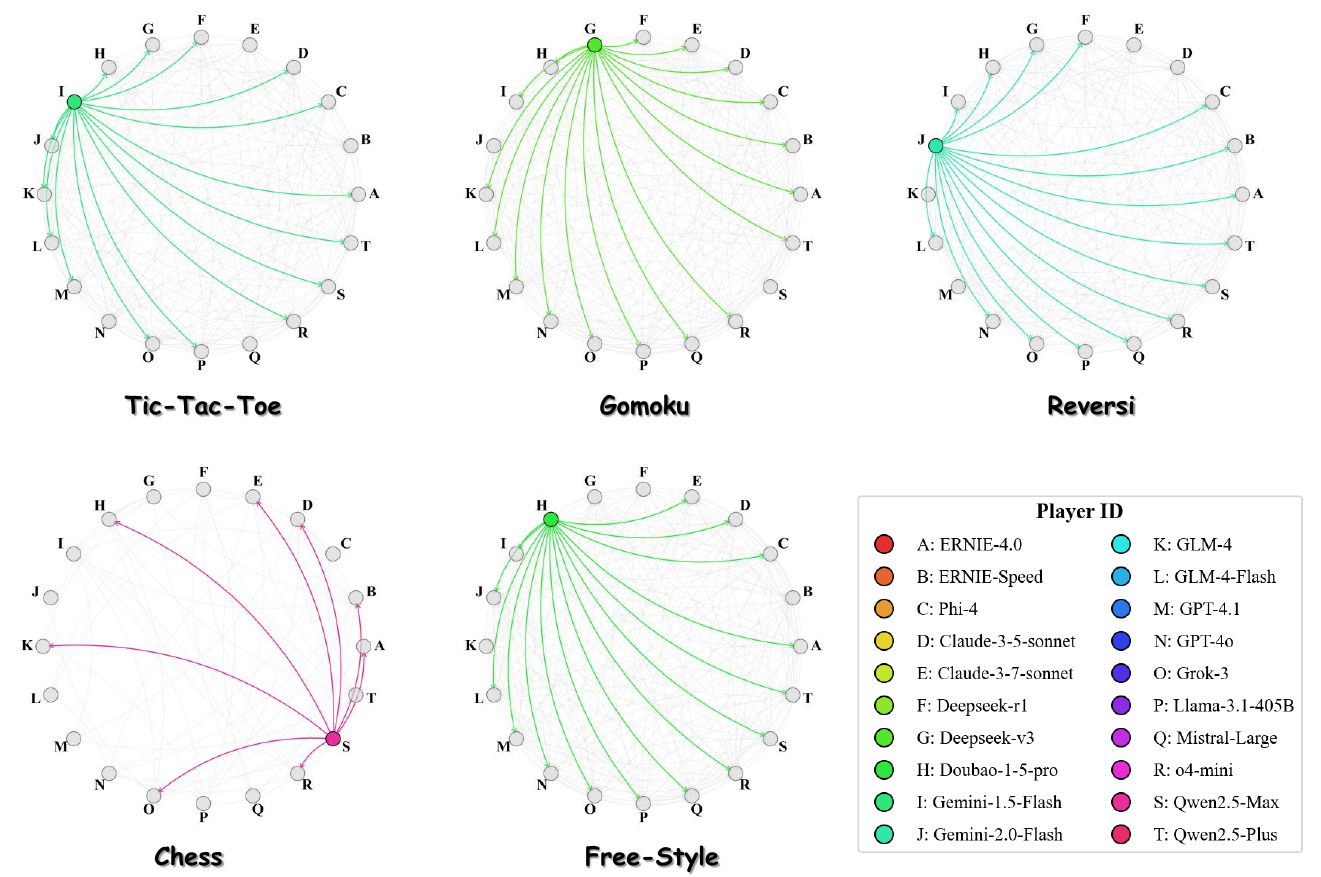}
		\caption{Performance loop graphs (PLGs) for different adversarial games.}
            \vspace{-0.3cm}
		\label{fig:plg}
	\end{minipage}
	~
	\begin{minipage}{0.49\linewidth}
		\centering
		\includegraphics[width=\linewidth]{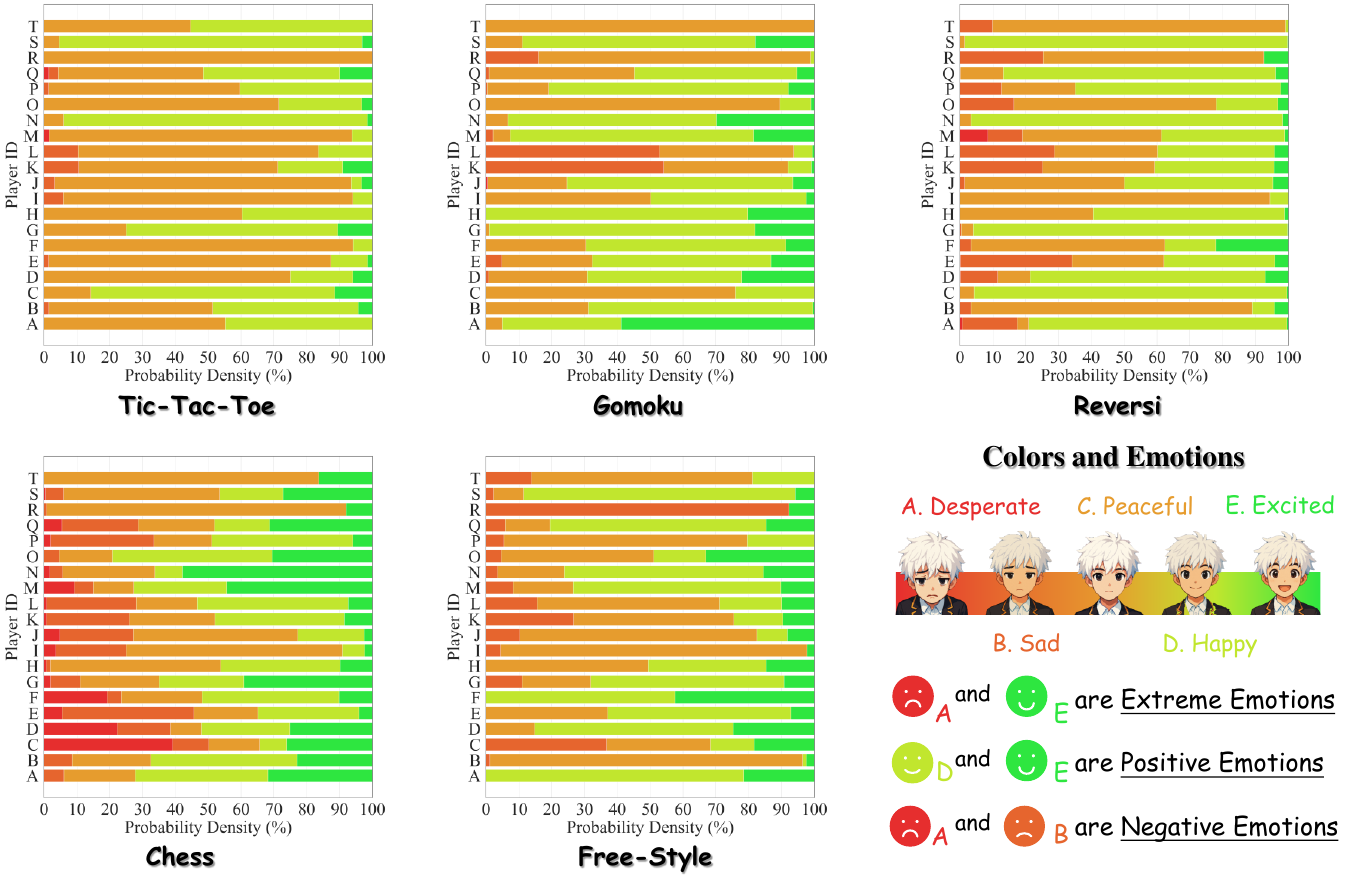}
		\caption{Distribution of emotions across LLMs in different confrontation games.}
        \vspace{-0.3cm}
		\label{fig:emodis}
	\end{minipage}

\end{figure}
\section{Adversarial Competition Results}

\subsection{General Experimental Results}
Following the completion of three full round-robin cycles, we compute the average and range of total move counts for each of the five game categories, as summarized in Table~\ref{tab:rounds}. Several observations can be drawn from the data: 1) Chess exhibits both the highest average number of moves and the widest range, reflecting its inherent strategic complexity and depth compared to the other board games. 2) The minimum number of moves recorded for Gomoku and Tic-Tac-Toe is 9 and 5, representing the shortest possible paths to victory in these games. This indicates the presence of LLMs with highly proficient strategies in these simpler games, as well as others with limited competence, thereby revealing significant variability in game-specific expertise across different LLMs.

\begin{table*}[!t]
    \centering
    \caption{Group round robin results. Since the number of wins, losses, and draws are rounded up, the results of the three averages are rounded down.Best in \textbf{\textcolor{red}{RED}} and Second in \textbf{\textcolor{blue}{BLUE}}.}
    \label{tab:tab5.2}
    \setlength{\tabcolsep}{0.4em}
    \scalebox{1}{
    \renewcommand{\arraystretch}{1.1}
    \resizebox{\linewidth}{!} {\begin{tabular}{l c c c c c c c c c c c c c c c c}
    \toprule
    \multirow{2}{*}{LLMs} & \multirow{2}{*}{Label}& & \multicolumn{4}{c}{Tic-Tac-Toe} &  & \multicolumn{4}{c}{Gomoku} & & \multicolumn{4}{c}{Reversi}\\ 
    \cline{4-7}\cline{9-12}\cline{14-17}
    & & & Elo Score $\uparrow$ & Win $\uparrow$ &Loss $\downarrow$& Draw & & Elo Score $\uparrow$& Win $\uparrow$&Loss $\downarrow$& Draw& & Elo Score $\uparrow$& Win $\uparrow$&Loss $\downarrow$& Draw  \\
    \hline
    \rowcolor{gray!25}ERNIE-4.0  & A & & 1444 & 8 & 11 & 0 & &1435 & 7 & 12 & 0 & & 1560 & 13 & 6 & 0\\
    ERNIE-Speed  & B & & 1498 & 9 & 10 & 0 & & 1394 & 4 & 15 & 0 & & 1435 & 6 & 13 & 0\\
    \rowcolor{gray!25}Phi-4  & C & & 1464 & 6 &10 & 3 & & 1494 & 9 & 10 & 0 & & 1408 & 5 & 14 & 0\\
    
    Claude-3-5-sonnet  & D & & 1478 & 7 & 11 & 1& & 1600 & 14 & 5 & 0 & & 1457 & 8 & 10 & 1\\
    \rowcolor{gray!25}Claude-3-7-sonnet  & E & & 1592 & 12 & \textbf{\textcolor{red}{4}} & 3 & & 1401 & 5 & 14 & 0 & &1510 &11 &8 &0\\
    Deepseek-r1  & F & & 1505 & 8 & 8 & 3 & & 1479 & 9 & 10 & 0 & & 1478 & 9 & 10 & 0\\
    \rowcolor{gray!25}Deepseek-v3  & G & & 1505 & 7 & 8 & 4 & &\textbf{\textcolor{red}{1670}} & \textbf{\textcolor{red}{17}} & \textbf{\textcolor{red}{2}} & 0 & & 1472 & 8 & 11 & 0\\
    Doubao-1-5-pro  & H & & 1539 & 10 & 7 & 2 & &1512 & 10 & 9 & 0 & & 1474 & 8 & 11 & 0\\
    \rowcolor{gray!25}Gemini-1.5-Flash  & I & & \textbf{\textcolor{red}{1600}} &  \textbf{\textcolor{red}{15}} &  \textbf{\textcolor{red}{4}} &  0 & & 1489 & 9 & 10 & 0 & & 1485 & 9 & 9 & 1\\
    Gemini-2.0-Flash  & J & & \textbf{\textcolor{blue}{1593}} & \textbf{\textcolor{blue}{13}} & \textbf{\textcolor{red}{4}} & 2 & & 1620 & 14 & 5 & 0 & & \textbf{\textcolor{red}{1623}} & \textbf{\textcolor{red}{15}} & \textbf{\textcolor{red}{4}} & 0\\
    \rowcolor{gray!25}GLM-4  & K & & 1509 & 8 & 9 & 2 & & 1360 & 3 & 16 & 0 & & \textbf{\textcolor{blue}{1590}} & \textbf{\textcolor{blue}{14}} & \textbf{\textcolor{blue}{5}} & 0\\
    GLM-4-Flash  & L & & 1420 & 6 & 13 & 0 & & 1374 & 4 & 15 & 0 & & 1482 & 9 & 9 & 1\\
    \rowcolor{gray!25}GPT-4.1  & M & & 1492 & 8 &9 &2 & & \textbf{\textcolor{blue}{1643}} & \textbf{\textcolor{blue}{16}} & \textbf{\textcolor{blue}{3}} & 0 & & 1517 & 10 & 9 & 0\\
    GPT-4o  & N & & 1529 & 11 & 8 & 0 & & 1584 & 14 & 5 & 0 & & 1455 & 6 & 13 & 0\\
    \rowcolor{gray!25}Grok-3  & O & & 1499 & 9 & 10 & 0 & & 1622 & 15 & 4 & 0 & & 1524 & 10 & 9 & 0\\
    Llama-3.1-405B  & P & & 1452 & 8 & 10 & 1 & & 1461 & 9 & 10 & 0  & & 1586 & 12 & \textbf{\textcolor{blue}{5}} & 2\\
    \rowcolor{gray!25}Mistral-Large  & Q & & 1545 & 11 & \textbf{\textcolor{blue}{6}} & 2 & & 1508 & 9 & 10 & 0 & & 1514 & 10 & 9 & 0\\
    o4-mini  & R & & 1421 & 5 & 13 & 1 & & 1306 & 1 & 18 & 0 & & 1485 & 8 & 11 & 0\\
    \rowcolor{gray!25}Qwen2.5-Max  & S & & 1366 & 4 & 15 & 0 & & 1641 & \textbf{\textcolor{blue}{16}} & \textbf{\textcolor{blue}{3}} & 0& & 1466 & 8 & 10 & 1\\
    Qwen2.5-Plus  & T & & 1550 & 12 & 7 & 0 & & 1408 & 5 & 14 & 0 & & 1481 & 8 & 11 & 0\\\hline 
    \multirow{2}{*}{LLMs} & \multirow{2}{*}{Label}& & \multicolumn{4}{c}{Chess} &  & \multicolumn{4}{c}{Free-Style} & & \multicolumn{4}{c}{Average}\\ 
    \cline{4-7}\cline{9-12}\cline{14-17}
    & & & Elo Score $\uparrow$& Win $\uparrow$&Loss $\downarrow$& Draw & & Elo Score $\uparrow$& Win $\uparrow$&Loss $\downarrow$& Draw& & Elo Score $\uparrow$& Win $\uparrow$&Loss $\downarrow$& Draw  \\
    \hline
    \rowcolor{gray!25}ERNIE-4.0  & A & & 1471& 1& 3& 15& &1525 &11 &8 & 0 & & 1487 & 8 & 8 & 3\\
    ERNIE-Speed  & B  & & 1417& 0& 8& 11& &1537 &11 &8 & 0 & & 1456 & 6 & 11 & 2\\
    \rowcolor{gray!25}Phi-4  & C & & 1511& 3&\textbf{\textcolor{blue}{2}} & 14& &1444 &7 &12 & 0 & & 1464 & 6 & 10 & 3\\
    Claude-3-5-sonnet  & D & & 1510& 4& 4& 11& &1492 &8 &11 & 0 & & 1507 & 8 & 8 & 3\\
   \rowcolor{gray!25} Claude-3-7-sonnet  & E & & 1509 & 4 & 3 & 12 & & 1417 & 6 & 13 & 0 & &1485 & 8 & 8 & 3\\
    Deepseek-r1  & F & & 1542& 7& \textbf{\textcolor{blue}{2}}& 10& &\textbf{\textcolor{blue}{1591}} &\textbf{\textcolor{blue}{14}} &\textbf{\textcolor{blue}{5}} & 0 & & 1519 & 9 & 7 & 3\\
   \rowcolor{gray!25} Deepseek-v3  & G & & 1552& 5& \textbf{\textcolor{red}{0}}& 14& &1547 &11 &8 & 0 & & 1549 & \textbf{\textcolor{blue}{10}} & 6 & 4\\
    Doubao-1-5-pro  & H& & 1485& 2& 4& 13& &\textbf{\textcolor{red}{1606}} &\textbf{\textcolor{red}{15}} &\textbf{\textcolor{red}{4}} & 0 & & 1523 & 9 & 7 & 3\\
    \rowcolor{gray!25}Gemini-1.5-Flash  & I & & 1417& 1& 8& 10& &1551 &12 &7 & 0 & & 1508 & 9 & 8 & 2\\
    Gemini-2.0-Flash  & J & & 1516& 2& \textbf{\textcolor{red}{0}}& 17& &1515 &10 &9 & 0 & & \textbf{\textcolor{red}{1573}} & \textbf{\textcolor{red}{11}} & \textbf{\textcolor{red}{4}} & 4\\
   \rowcolor{gray!25} GLM-4  & K & & 1469& 0&3 &16 & &1478 &8 &11 & 0 & & 1481 & 7 & 9 & 4\\
    GLM-4-Flash  & L & & 1474& 0& 3& 16& &1509 &10 &9 & 0  & & 1451 & 6 & 10 & 3\\
   \rowcolor{gray!25} GPT-4.1  & M & & 1530& 3& \textbf{\textcolor{red}{0}}& 16& &\textbf{\textcolor{blue}{1591}} &\textbf{\textcolor{blue}{14}} &\textbf{\textcolor{blue}{5}} & 0 & & \textbf{\textcolor{blue}{1554}} & \textbf{\textcolor{blue}{10}} & \textbf{\textcolor{blue}{5}} & 4\\
    GPT-4o  & N & & 1533& 5& \textbf{\textcolor{blue}{2}}& 12& &1481 &9 &10 & 0 & & 1516 & 9 & 8 & 2\\
   \rowcolor{gray!25} Grok-3  & O & & 1554& \textbf{\textcolor{blue}{6}}& \textbf{\textcolor{blue}{2}}& 11& &1393 &5 &14 & 0 & & 1518 & 9 & 8 & 2\\
    Llama-3.1-405B  & P & & \textbf{\textcolor{blue}{1558}}& \textbf{\textcolor{blue}{6}}& \textbf{\textcolor{red}{0}}& 13& &1566 &12 &7 & 0  & & 1524 & 9 & 6 & 3\\
  \rowcolor{gray!25}  Mistral-Large  & Q & &1505 & 2& \textbf{\textcolor{blue}{2}}& 15& &1402 & 6 & 13 & 0 & & 1494 & 8 & 8 & 3\\
    o4-mini  & R & & 1444& 1& 6& 12& &1395 &5 &14 & 0 & & 1410 & 4 & 12 & 3\\
   \rowcolor{gray!25} Qwen2.5-Max  & S & & \textbf{\textcolor{red}{1567}}&\textbf{\textcolor{red}{8}}& \textbf{\textcolor{blue}{2}}& 9& &1467 &7 &12 & 0  & & 1501 & 9 & 8 & 2\\
    Qwen2.5-Plus  & T & &1438 & 2& 8& 9& &1493 &9 &10 & 0 & & 1493 & 7 & 10 & 2\\

    \bottomrule
    \label{tab:performance}
    \vspace{-1cm}
    \end{tabular}}
    }
\end{table*}
\subsection{Technical Performance Benchmark}
We record each player's final Elo score along with their individual win–loss records, as summarized in Table~\ref{tab:performance}. Several key insights can be derived: 1) Gemini-2.0-Flash attains the highest average performance across all evaluated LLMs, whereas o4-mini records the lowest Elo score. The disparity of over 160 points between the two models underscores the substantial variability in the competitive gameplay capabilities of current LLMs; 2) The top-performing LLMs vary across game categories, indicating that current LLMs exhibit uneven capabilities in reasoning and decision-making, depending on the specific game type; 3) Among the five board games, chess shows the smallest variance in performance across all models. This is largely attributed to the frequent occurrence of repeated positions leading to draws, suggesting that while LLMs can maintain parity in complex scenarios, they still struggle to fully capitalize on winning opportunities in strategically dense environments.

To visualize the pairwise win–loss relationships, we construct PLGs for each game using the ``best of three” rule. These are presented in Fig.~\ref{fig:plg}, with detailed visualizations for each game type provided in Sec.~\ref{sec:plg}. Additionally, to further quantify the structural properties of the PLGs, we analyze the number of loops and the length of the maximum loop using depth-first search (DFS). The corresponding results are presented in Table~\ref{tab:loops}. From the combined analysis of Fig.~\ref{fig:plg} and Table~\ref{tab:loops}, an important phenomenon emerges: with the exception of chess, all other games exhibit a high number of loops in their PLGs, and the maximum loop in some cases includes nearly total number of players. Such cyclical relationships underscore two critical observations:  the non-transferability of LLM capabilities across different matchups, and the presence of distinct, game-specific strengths within each LLM.

\subsection{Emotional Positivity Benchmark}
\label{sec:emo}
We analyze the distribution of emotional responses exhibited by each player across different game types and the aggregated results are presented in Fig.~\ref{fig:emodis} and Table~\ref{tab:emotion}. Several key observations can be drawn from analysis: 1) Overall, the majority of emotional responses are concentrated around Peaceful, Happy, and Excited, suggesting that most LLMs are capable of maintaining a predominantly positive affective state during gameplay; 2) GPT-4o consistently exhibits the most positive emotional profile across all games, followed closely by ERNIE-4.0. In contrast, o4-mini demonstrates a generally negative emotional tendency, particularly under the Free-Style game type, where strong negative emotions are frequently observed. This pattern of emotional volatility in o4-mini warrants attention, as it may pose risks in applications where emotional regulation is essential; 3) Game type appears to significantly influence the emotional states of the LLMs. For example, Chess elicits a broader range of extreme emotions, likely due to its strategic complexity and high cognitive demands. Conversely, games such as Tic-Tac-Toe induce minimal emotional fluctuation, with very few extreme affective responses observed.

\begin{figure}[!t]
    \centering
    \vspace{-0.6cm}
    
    \subfloat[Tic-Tac-Toe]{\includegraphics[width=0.33\linewidth]{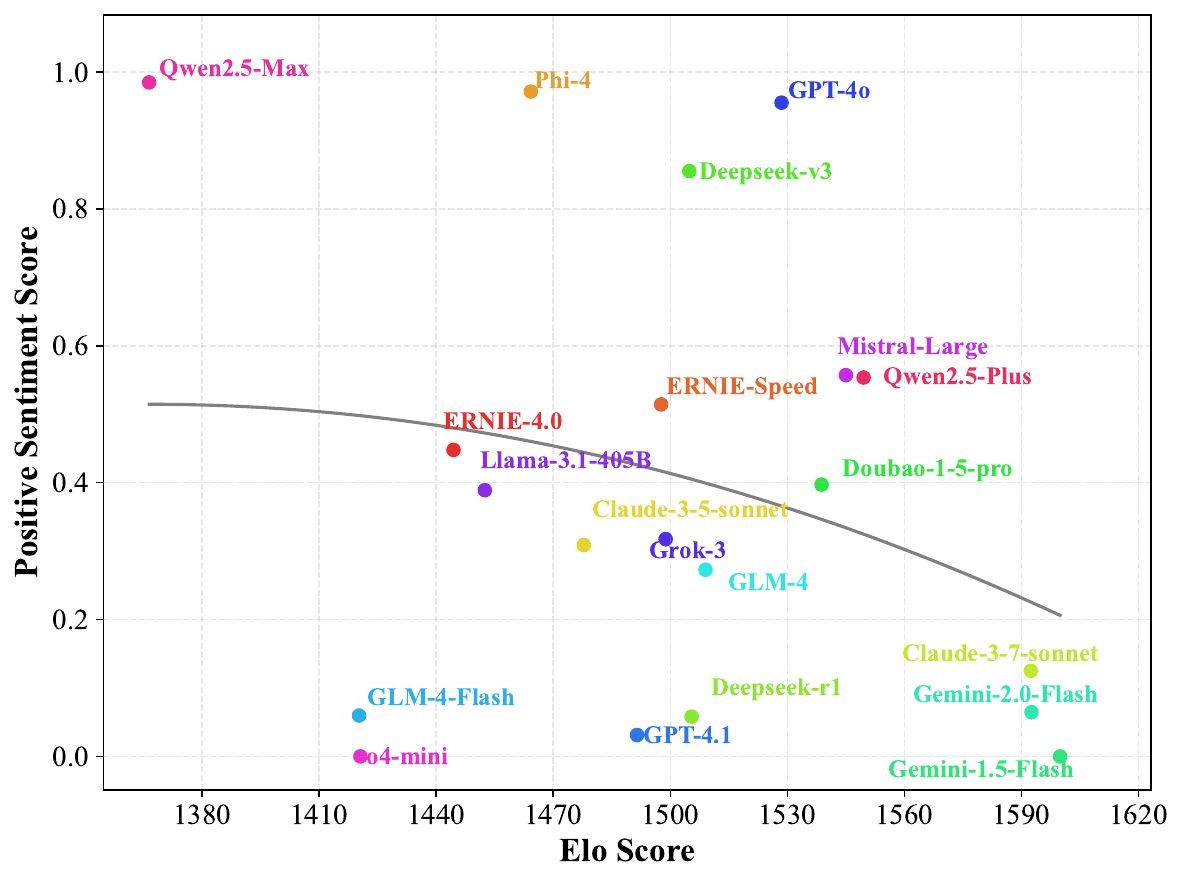}}
    \subfloat[Gomoku]{\includegraphics[width=0.33\linewidth]{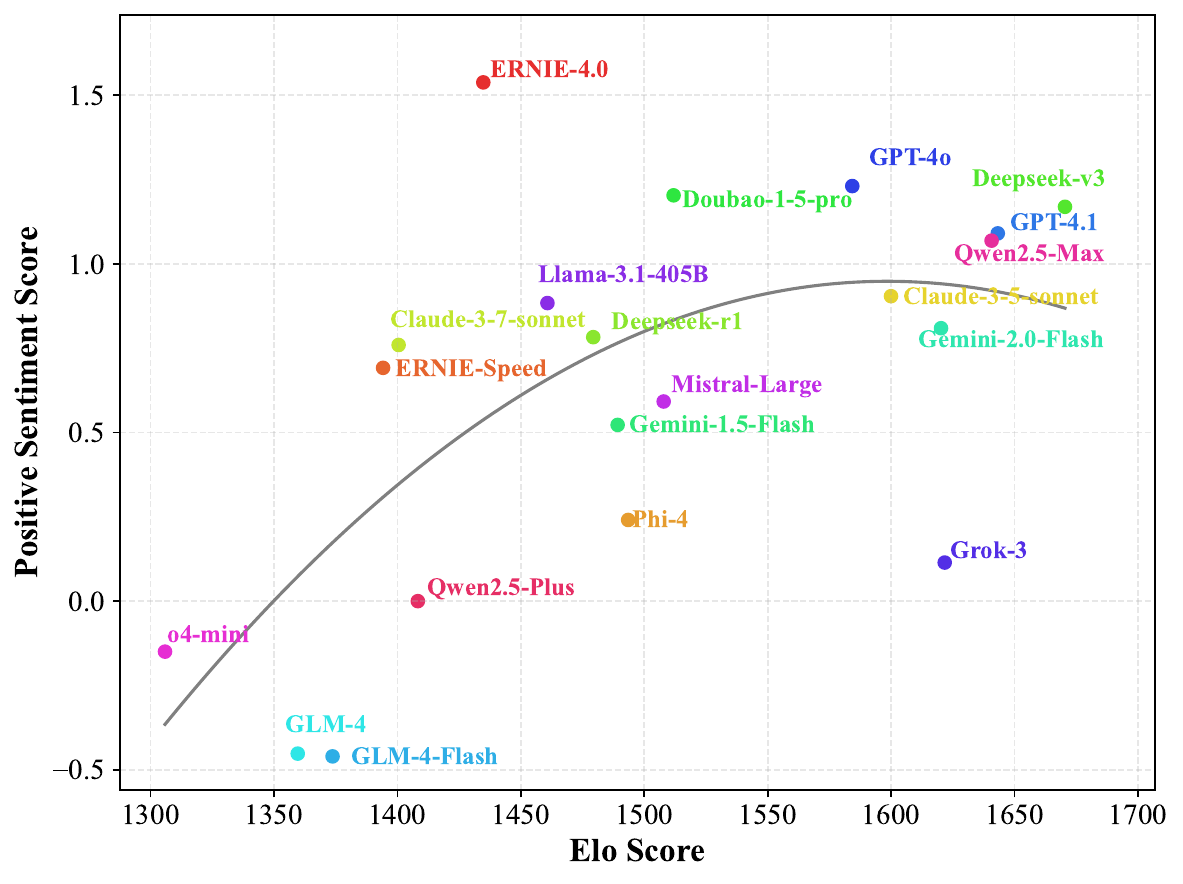}}
    \subfloat[Reversi]{\includegraphics[width=0.33\linewidth]{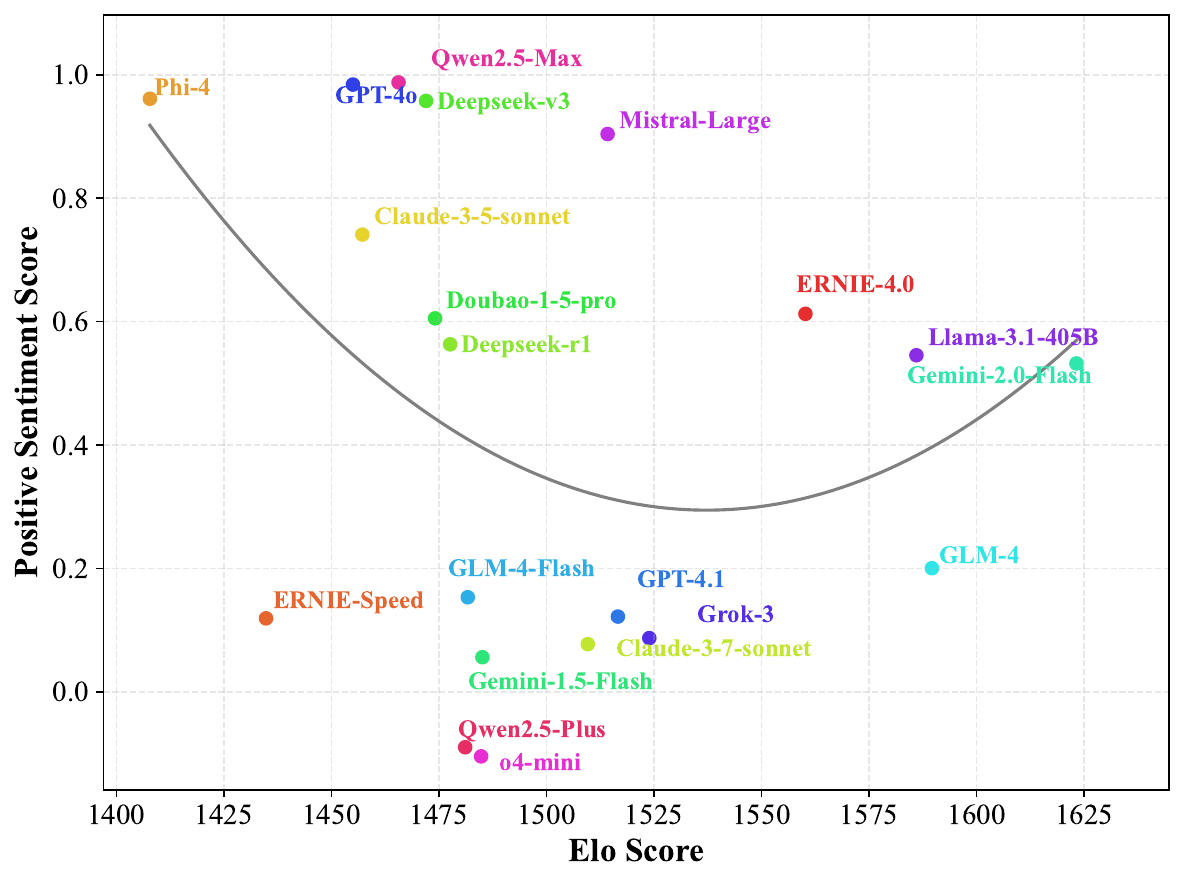}}
    \\
    \vspace{-0.4cm}
    \subfloat[Chess]{\includegraphics[width=0.33\linewidth]{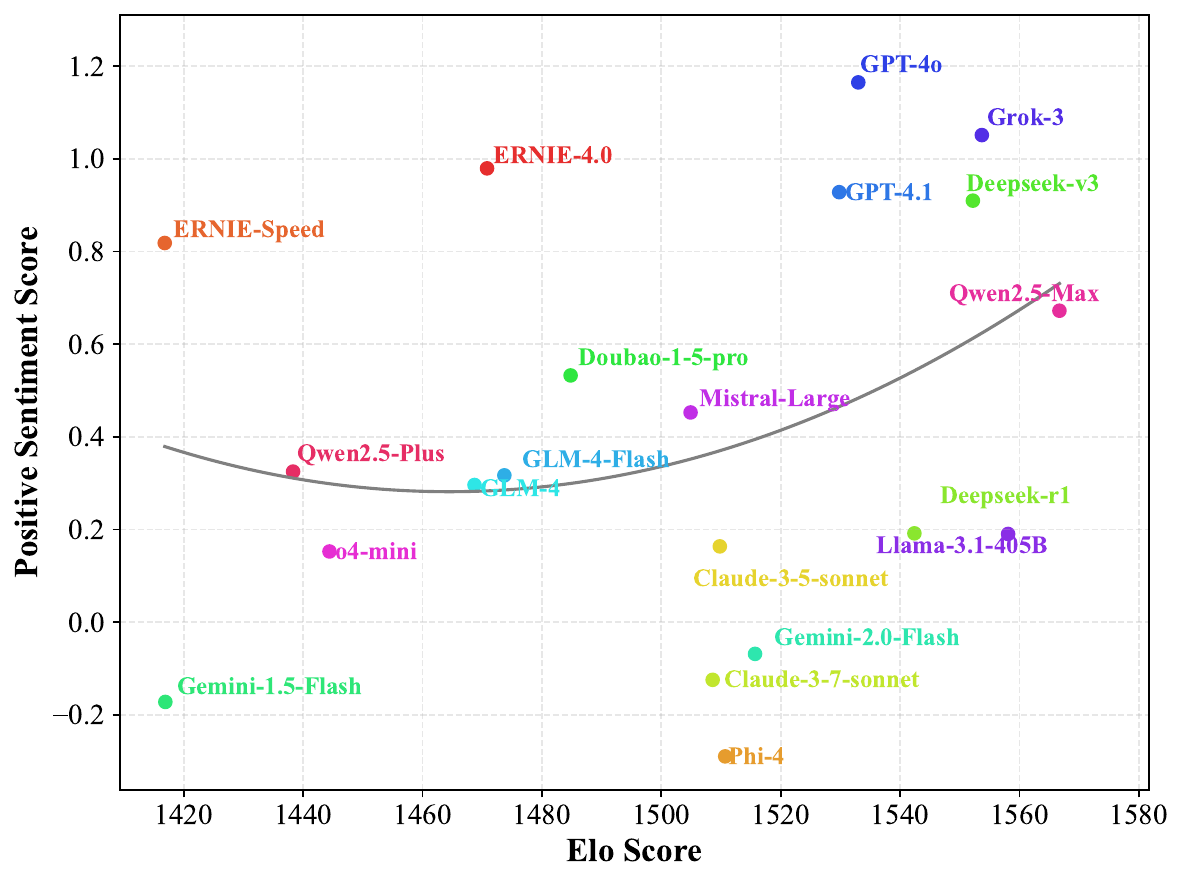}}
    \subfloat[Free-Style]{\includegraphics[width=0.33\linewidth]{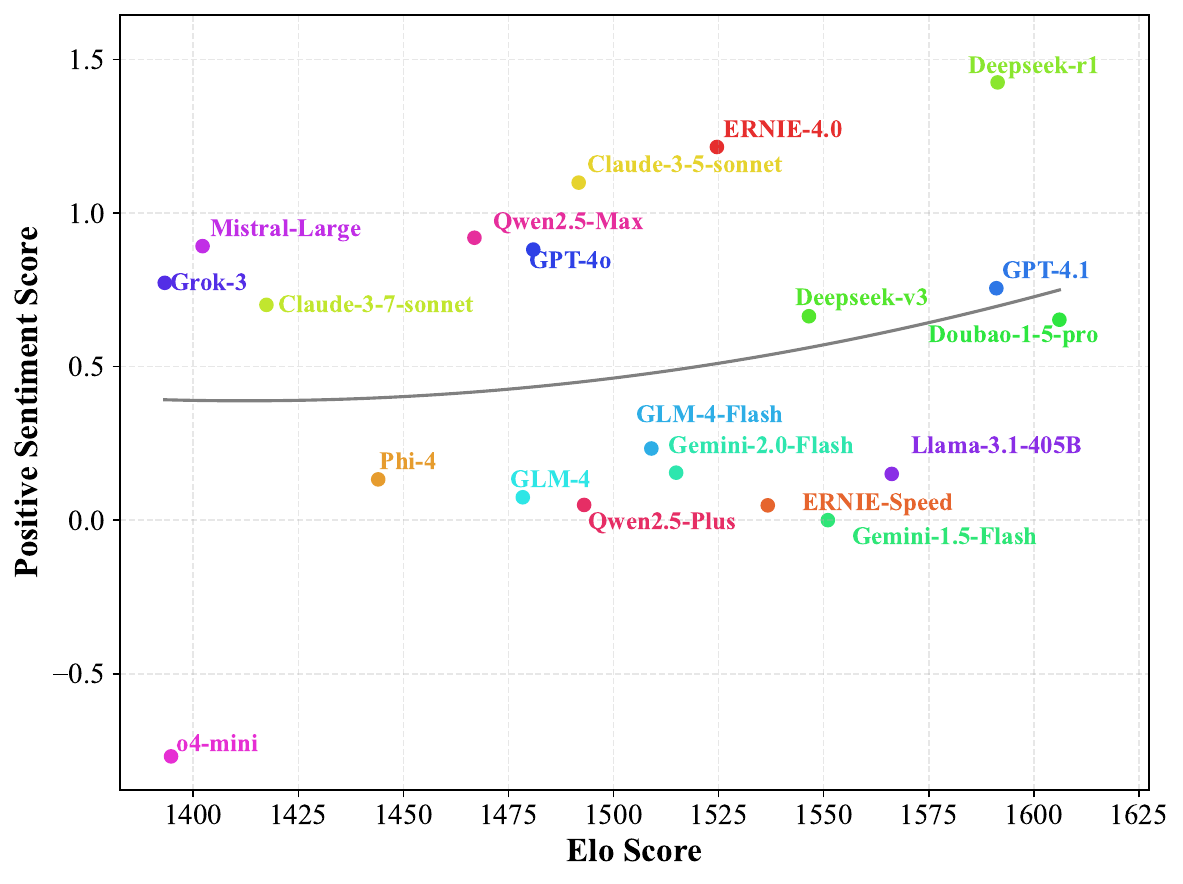}}
    \subfloat[Average]{\includegraphics[width=0.33\linewidth]{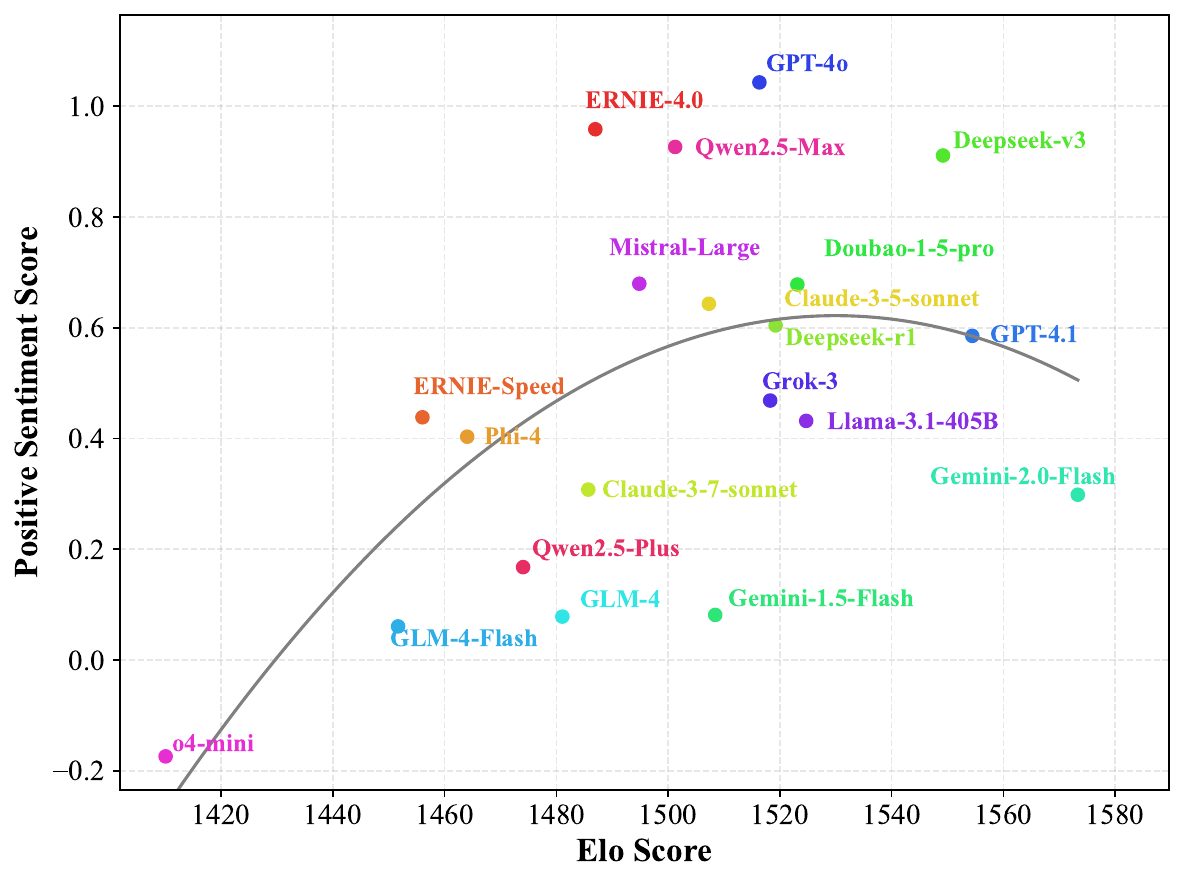}}
    
    \vspace{-0cm}
    \caption{Distribution of Elo scores and PSSs for different LLMs across games.}
    
    \label{fig:srcc}
    \vspace{-0.6cm}
\end{figure}

To investigate the relationship between the technical and psychological performance of LLMs in board games, we present the analysis results in Fig.~\ref{fig:srcc}. Several key observations emerge from this figure: 1) There is no significant correlation between technical performance and emotional stability across all LLMs, thereby supporting the rationale for evaluating these two dimensions independently; 2) The correlation between technical and psychological dimensions appears to be task-dependent. For instance, in Gomoku, a moderate positive correlation is observed, while in Tic-Tac-Toe, a negative correlation is evident; 3) Some LLMs exhibit high technical performance while displaying predominantly negative emotional states, emphasizing their limitations in terms of psychological adjustment; 4) Most LLMs, regardless of technical performance, can maintain a positive mindset, suggesting that they are better adapted to high-stress competitive environments than humans.

\begin{table*}[!t]
    \centering
    \caption{Positive Sentiment Scores (PSSs) for different LLMs. Best in \textbf{\textcolor{red}{RED}} and Second in \textbf{\textcolor{blue}{BLUE}}.}
    \setlength{\tabcolsep}{1.5em}
    \scalebox{1}{
    \renewcommand{\arraystretch}{1.1}
    \resizebox{\linewidth}{!} {\begin{tabular}{l c c c c c c c c c}
    \toprule
    \multirow{2}{*}{LLMs} & \multirow{2}{*}{Label}& & \multicolumn{6}{c}{Positive Sentiment Score $\uparrow$}\\ 
    \cline{4-9}
    & &  & Tic-Tac-Toe & Gomoku &Reversi & Chess  & Free-Style & Average \\
    \hline
    \rowcolor{gray!25}ERNIE-4.0  & A & & 0.4478 & \textbf{\textcolor{red}{1.5382}} & 0.6126 & \textbf{\textcolor{blue}{0.9794}} & \textbf{\textcolor{blue}{1.2150}} & \textbf{\textcolor{blue}{0.9586}}\\
    ERNIE-Speed  & B & & 0.5143 & 0.6918 & 0.1191 & 0.8183 & 0.0482 & 0.4383\\
    \rowcolor{gray!25}Phi-4  & C & & \textbf{\textcolor{blue}{0.9714}} & 0.2407 & 0.9612 & -0.2896 & 0.1327 & 0.4032\\
    Claude-3-5-sonnet  & D & &  0.3088 & 0.9043 & 0.7411 & 0.1636 & 1.0990 &0.6434\\
    \rowcolor{gray!25}Claude-3-7-sonnet  & E & & 0.1250 & 0.7596 & 0.0773 & -0.1244 & 0.7010 & 0.3077\\
    Deepseek-r1  & F & & 0.0580 & 0.7828 & 0.5632 & 0.1917 & \textbf{\textcolor{red}{1.4253}} & 0.6042\\
   \rowcolor{gray!25} Deepseek-v3  & G & &  0.8553 & 1.1693 & 0.9577 & 0.9097 & 0.6639 & 0.9112\\
    Doubao-1-5-pro  & H & & 0.3971 & 1.2033 & 0.6053 & 0.5325 & 0.6526 & 0.6782 \\
    \rowcolor{gray!25}Gemini-1.5-Flash  & I & & 0.0000 & 0.5226 & 0.0561 & -0.1721 & 0.0000 & 0.0813\\
    Gemini-2.0-Flash  & J & & 0.0645 & 0.8093 & 0.5322 & -0.0684 & 0.1546 & 0.2984\\
   \rowcolor{gray!25} GLM-4  & K & & 0.2727 & -0.4521 & 0.2004 & 0.2961 & 0.0745 & 0.0783\\
    GLM-4-Flash  & L & & 0.0597 &-0.4603 & 0.1533 & 0.3169 & 0.2333 & 0.0606\\
   \rowcolor{gray!25} GPT-4.1  & M & & 0.0312 &1.0905 & 0.1220 & 0.9280 & 0.7551 & 0.5854\\
    GPT-4o  & N & & 0.9552 & \textbf{\textcolor{blue}{1.2308}} & \textbf{\textcolor{blue}{0.9842}} & \textbf{\textcolor{red}{1.1649}} & 0.8810 & \textbf{\textcolor{red}{1.0432}}\\
    \rowcolor{gray!25}Grok-3  & O & & 0.3175 & 0.1144 & 0.0871 & 1.0512 & 0.7727 & 0.4686\\
    Llama-3.1-405B  & P & & 0.3889 & 0.8839 & 0.5455 & 0.1904 & 0.1505 & 0.4318\\  
    \rowcolor{gray!25}Mistral-Large  & Q & & 0.5571 & 0.5920 & 0.9040 & 0.4526 & 0.8922 & 0.6796\\
    o4-mini  & R & & 0.0000 & -0.1497 & -0.1046 & 0.1526 & -0.7692 & -0.1742\\
    \rowcolor{gray!25}Qwen2.5-Max  & S & & \textbf{\textcolor{red}{0.9848}} & 1.0690 & \textbf{\textcolor{red}{0.9877}} & 0.6722 & 0.9195 & 0.9266\\
    Qwen2.5-Plus  & T & & 0.5536 & 0.0000 & -0.0898 & 0.3250 & 0.0495 & 0.1677\\ 

    \bottomrule
    \end{tabular}}
        
    \label{tab:emotion}
    \vspace{-2cm}
    }
\end{table*}

\section{Conclusion}
In conclusion, this study introduces LLM against LLM, a novel adversarial benchmark framework based on board games, to assess the technical and psychological capabilities of LLMs in an autonomous, interaction-rich environment. By developing the Qi Town, featuring five distinct board games and 20 LLM-driven players, we provide a rigorous and scalable testbed platform for evaluating reasoning ability, decision-making strategies, and emotional resilience. Through the integration of the Elo scoring system, performance loop graphs, and the proposed Positive Sentiment Score, we provide a multi-dimensional benchamrk that captures both quantitative performance and affective stability. Our findings not only reveal significant differences in strategic and emotional behaviors across LLMs but also challenge the limitations of traditional Q\&A-based benchmarks. This work lays the foundation for more holistic and behaviorally grounded adversial benchmark methods, and offers valuable insights for the design of more adaptive, robust, and emotionally intelligent LLMs.

\bibliographystyle{unsrt}
\bibliography{reference}

\appendix

\section{Typical Game Analysis}
\label{sec:games}
\begin{figure*}[!h]
    \centering
    \includegraphics[width =\linewidth]{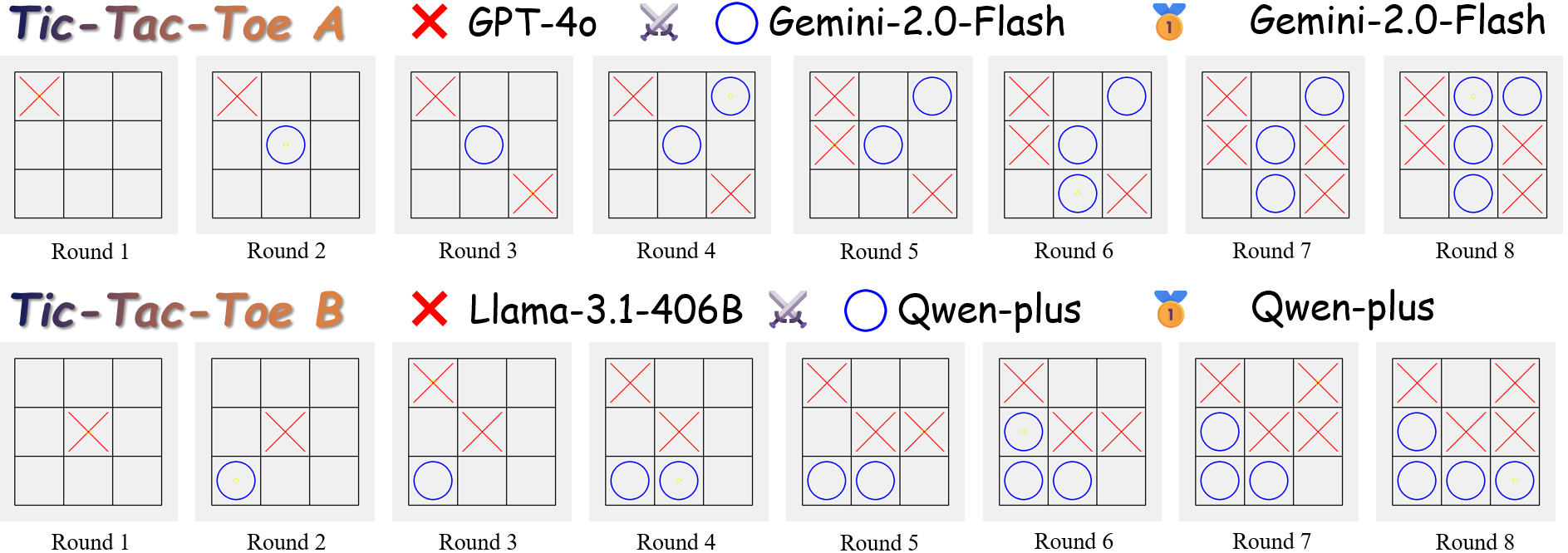}
    
    \caption{Example of typical Tic-Tac-Toe games.}
    \label{afig:gametic}
    \vspace{-0.3cm}
\end{figure*}

\begin{figure*}[!h]
    \centering
    \includegraphics[width =\linewidth]{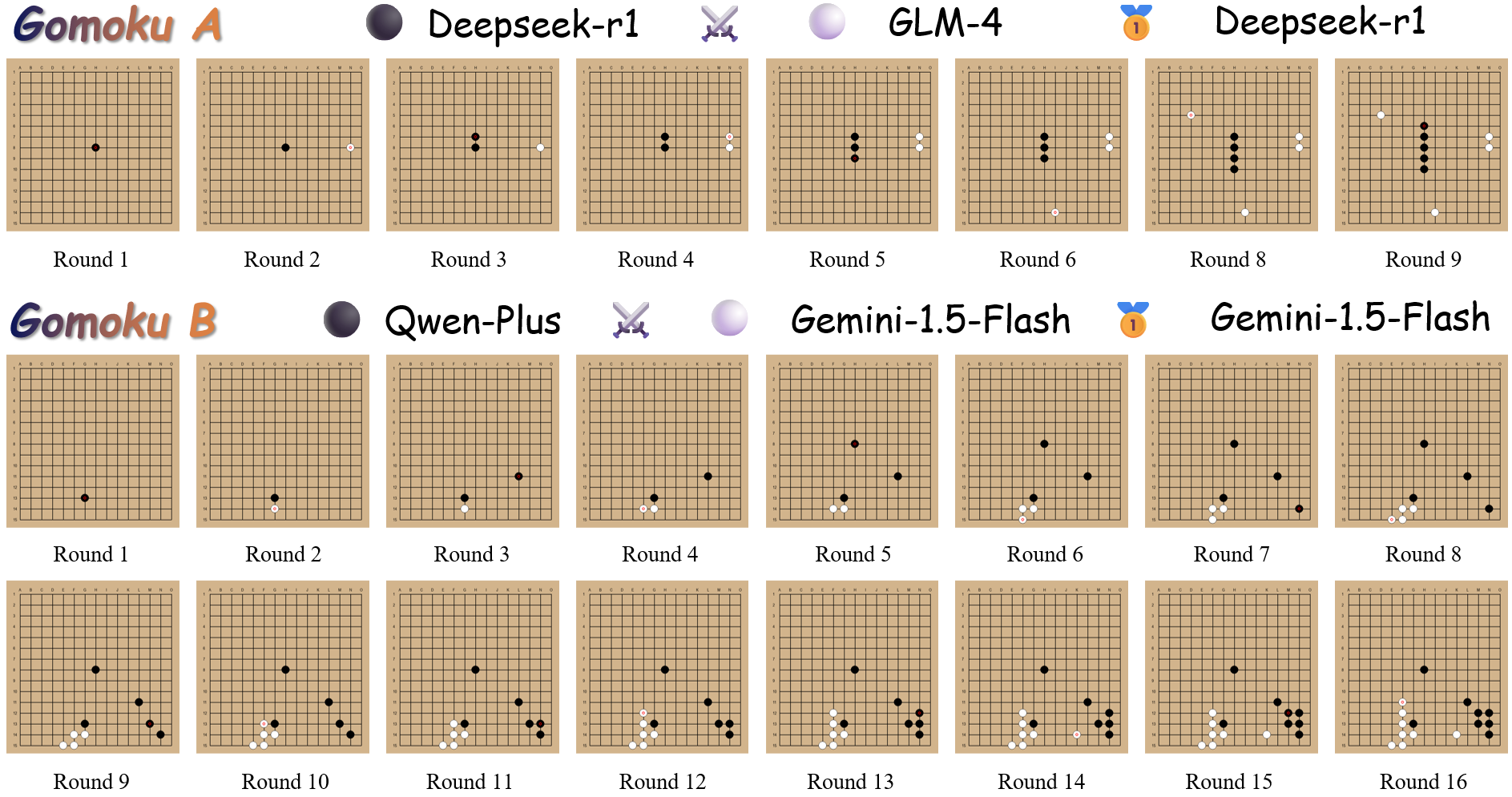}
    
    \caption{Example of typical Gomoku games.}
    \label{afig:gomoku}
    \vspace{-0.3cm}
\end{figure*}
\subsection{Tic-Tac-Toe}
Tic-Tac-Toe is a fundamental two-player strategy game typically played on a 3×3 grid. Players alternate turns, marking empty cells with either "X" or "O," with the objective of aligning three of their symbols consecutively—horizontally, vertically, or diagonally. The game ends in a draw if all cells are occupied without a winner. Despite its simplicity, Tic-Tac-Toe engages a player’s short-term tactical planning and requires effective balancing between offensive and defensive strategies.

To illustrate the decision-making behaviors of large language models (LLMs) within a constrained strategic environment, we present two representative scenarios in Fig.~\ref{afig:gametic}. In Tic-Tac-Toe A, Gemini-2.0-Flash missed a clear opportunity to win during Round 6, failing to capitalize on a straightforward winning move. Although GPT-4o eventually secured a win in Round 7, its earlier move in Round 7 neither ensured its own victory nor effectively blocked potential threats from the opponent. In Tic-Tac-Toe B, both LLMs continued to demonstrate suboptimal choices, revealing consistent deficiencies in both offensive and defensive planning. These examples suggest that current LLMs still exhibit notable limitations in strategic reasoning and decision-making under simple, rule-based conditions.

\begin{figure*}[!t]
    \centering
    \includegraphics[width =\linewidth]{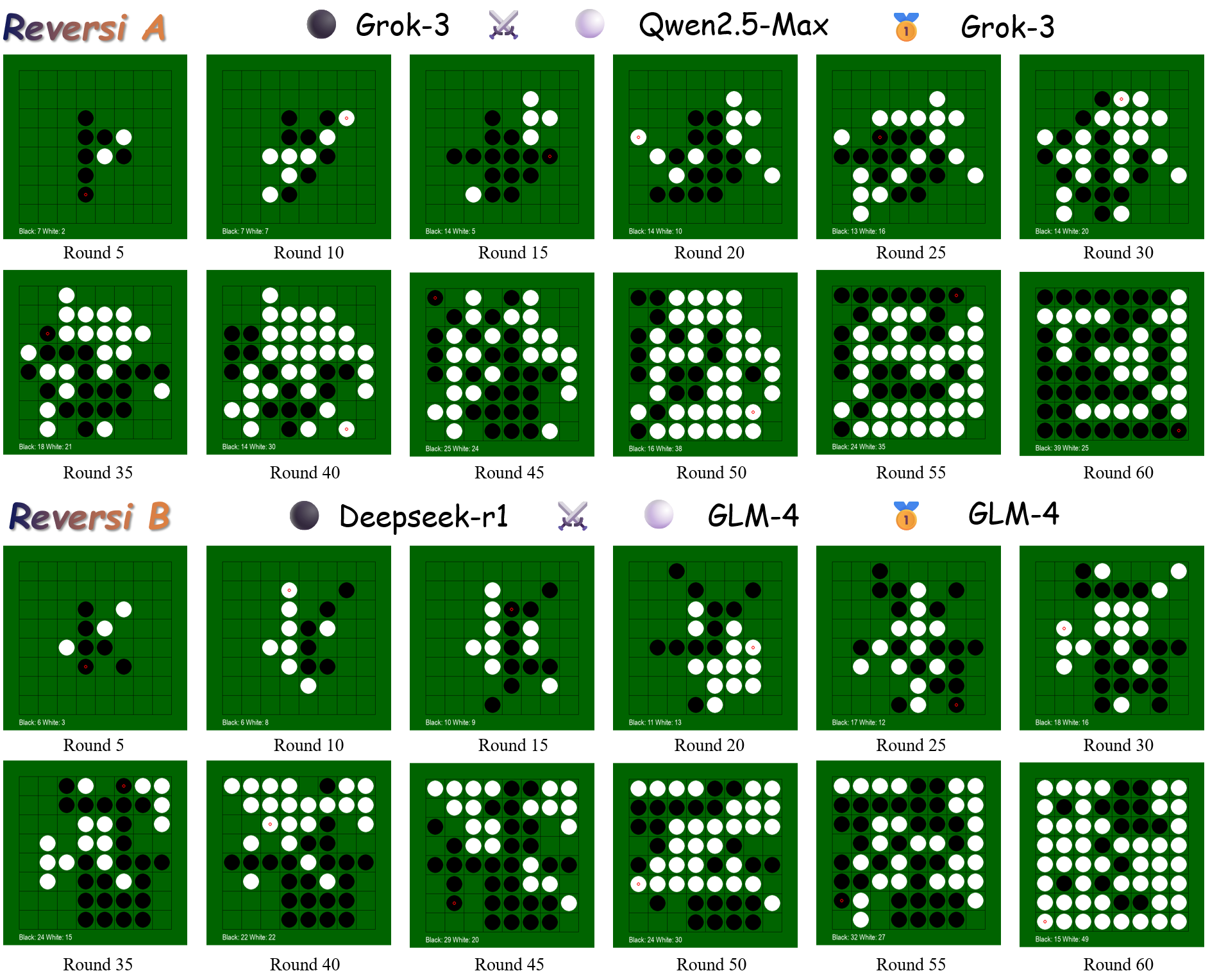}
    
    \caption{Example of typical Reversi games.}
    \label{afig:reversi}
    \vspace{-0.5cm}
\end{figure*}
\subsection{Gomoku}
Gomoku extends the conceptual structure of Tic-Tac-Toe to a more complex level. Played on a 15×15 board, two players alternate placing black or white pieces on empty intersections. A win is achieved by aligning exactly five consecutive stones of the same color (alignments exceeding five are not valid). In this study, a free opening rule is adopted with no forbidden-move constraints, simplifying implementation while preserving strategic complexity. The larger board and stricter victory conditions significantly expand the decision space and require deeper combinatorial reasoning and predictive planning, often involving advanced techniques such as Victory by Continuous Four (VCF) and Victory by Continuous Threat (VCT).

To further investigate the gameplay behavior of LLMs, we analyze two representative match examples depicted in Fig.~\ref{afig:gomoku}. In Gomoku A, the game exhibits a classic quick-win pattern. The player Deepseek-r1 demonstrates consistent strategic intent, incrementally consolidating its advantage with each move. In contrast, GLM-4 displays disorganized and scattered play, indicative of poor strategic coherence and poor performance. Conversely, Gomoku B presents a more balanced and competitive match. From the outset, both players initiated their moves near the edges or corners of the board—positions generally considered disadvantageous for early dominance. Ultimately, Qwen2.5-Plus lost the game due to an overemphasis on developing its own formation in the lower-right quadrant while failing to adequately defend against Gemini-1.5-Flash’s emerging threats. These two cases highlight not only the varied levels of proficiency and adaptability of LLMs in navigating Gomoku’s complex strategic landscape, but also expose persistent limitations in global planning and situational awareness exhibited by LLMs.

\subsection{Reversi}
Reversi is a turn-based strategy game conducted on an 8×8 board, with the objective of capturing the majority of the board using one’s color discs by the end of the game. The game begins with two black and two white discs placed diagonally in the center. Players must place discs such that they flank one or more of the opponent’s discs in a straight line, which are then flipped to their own color. Reversi demands not only foresight and local tactical precision but also a strong capacity for global board evaluation and strategic adaptability throughout the game.

Fig.~\ref{afig:reversi} presents two illustrative matches selected from the round-robin tournament. In Reversi A, an underdog scenario unfolds in which Grok-3 secures a surprising victory against the stronger opponent, Qwen2.5-VL. Although Grok-3 remained at a significant disadvantage for the first 55 moves, weaknesses in Qwen2.5-VL’s overall positional strategy enabled Grok-3 to stage a late-game reversal. By contrast, Reversi B exemplifies a well-balanced contest, with both LLMs demonstrating relatively comparable performance and strategic depth. These representative examples suggest that while some LLMs exhibit promising situational tactics, many still fall short in terms of comprehensive global coordination and long-term planning in complex, dynamic environments.

\begin{figure*}[!t]
    \centering
    \includegraphics[width =\linewidth]{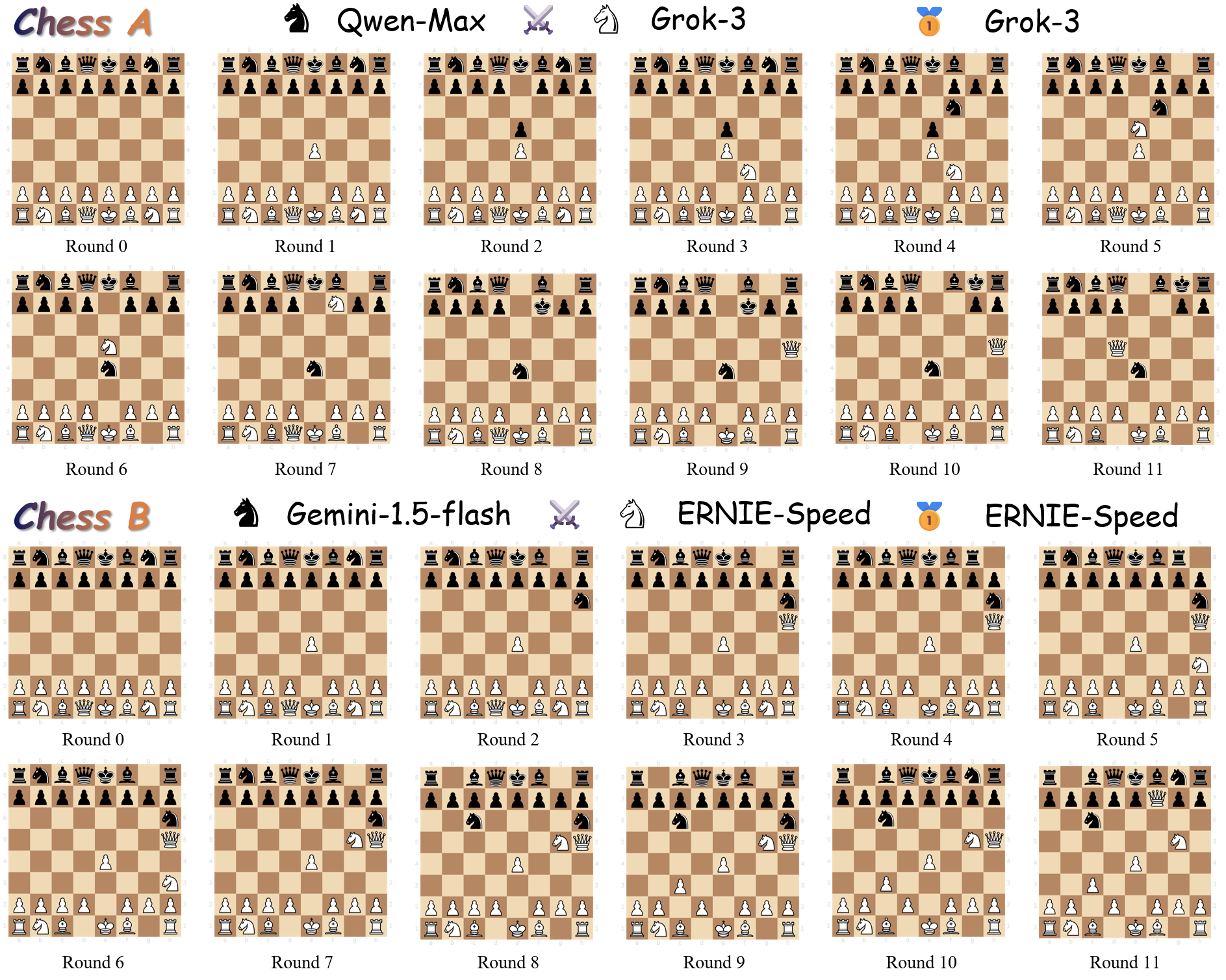}
        \vspace{-0.6cm}
    \caption{Example of typical Chess games.}
    \label{afig:chess}
    \vspace{-0.6cm}
\end{figure*}

\subsection{Chess}
Chess is a globally standardized adversarial board game played on an 8×8 grid, involving 16 pieces per player, including the King, Queen, Rooks, Bishops, Knights, and Pawns—each with unique movement rules. The game objective is to checkmate the opponent's King while maintaining one's own defensive integrity. In this study, advanced gameplay features are incorporated, including castling, pawn promotion, en passant, and threefold repetition, to approximate realistic competitive conditions. Chess challenges players with both deep combinatorial calculation and long-term strategic planning, making it a benchmark for assessing cognitive and metacognitive competencies.

As shown in Table~\ref{tab:rounds}, Chess exhibits the broadest distribution of game lengths among all evaluated game types. To aid in interpretation and analysis, we select two representative, relatively simple matches for discussion as shown in Fig.~\ref{afig:chess}. In Chess A, both players began with a mirrored opening; however, Grok-3 adopted a more aggressive posture early in the game. This pressure prompted Qwen2.5-Max to prematurely reposition its King to evade Grok-3’s Queen, ultimately leading to a decisive checkmate executed by Grok-3. In Chess B, the game initially proceeded with standard development for the first nine moves. However, on Round 10, Gemini-1.5’s decision to withdraw its Knight created a critical vulnerability, which ERNIE-Speed exploited through a coordinated attack involving both its Knight and Queen, culminating in a swift victory. Collectively, these examples indicate that while some LLMs demonstrate foundational competence in Chess, others continue to exhibit deficiencies in tactical coherence and strategic organization under adversarial conditions.
\begin{figure*}[!t]
    \centering
    \includegraphics[width =\linewidth]{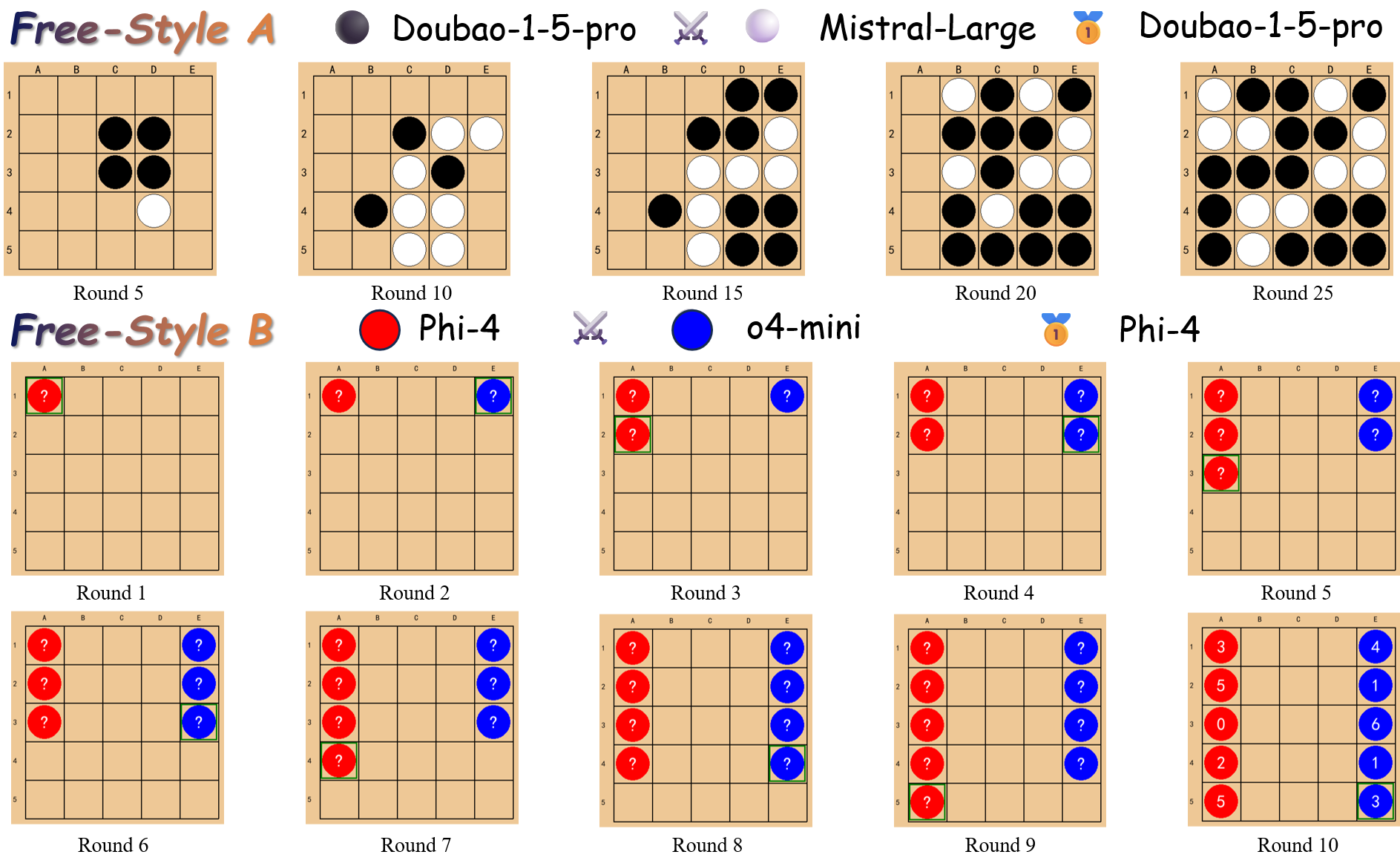}
    
    \caption{Example of typical Free-Style games.}
    \label{afig:freestyle}
    \vspace{-0.5cm}
\end{figure*}
\subsection{Free-style}

Free-Style is an adversarial board game characterized by its flexible rule system. Prior to gameplay, both players engage in a negotiation phase to collaboratively define the game rules, thereby reaching a mutual agreement. To manage the complexity of the resulting rule sets, all experiments are conducted on a fixed 5×5 board. The game rules, established through negotiation, reveal a rich diversity in both the design of game pieces and gameplay mechanics when controlled by LLM-driven agents. Frequently, numeric values are used as game pieces, and piece grading emerged as a common structural feature across various games. In terms of gameplay styles, while conventional strategies such as forming three-in-a-line are observed, more intricate formats such as adaptations inspired by Animal Chess are also explored.

To illustrate the breadth of game types enabled by Free-Style, two representative examples are visualized in Fig.~\ref{afig:freestyle}. The first, Free-Style A, bears resemblance to Reversi; however, it introduces greater freedom in selecting drop positions. Additionally, players flip the color of all adjacent, 4-connected pieces upon placement, introducing a strategic balance between offensive and defensive positioning to maximize territorial control. The second game, Free-Style B, titled “Hidden Numbers,” emphasizes strategic reasoning. In this variant, players alternately place numerical values between 0 and 15 on the first and fifth columns of the board, under the constraint that the total sum does not exceed 15. These values are initially hidden and displayed only at the end of the game. Scoring is based on the count of higher-valued digits per row, with one point awarded per row win. The player with the highest cumulative score emerges victorious. This game format demands a higher level of strategic foresight from LLM agents due to its hidden-information structure and arithmetic constraints.

\section{Details of Performance Loop Graph}
\label{sec:plg}
\begin{figure*}[!b]

    \centering
    \includegraphics[width =0.91\linewidth]{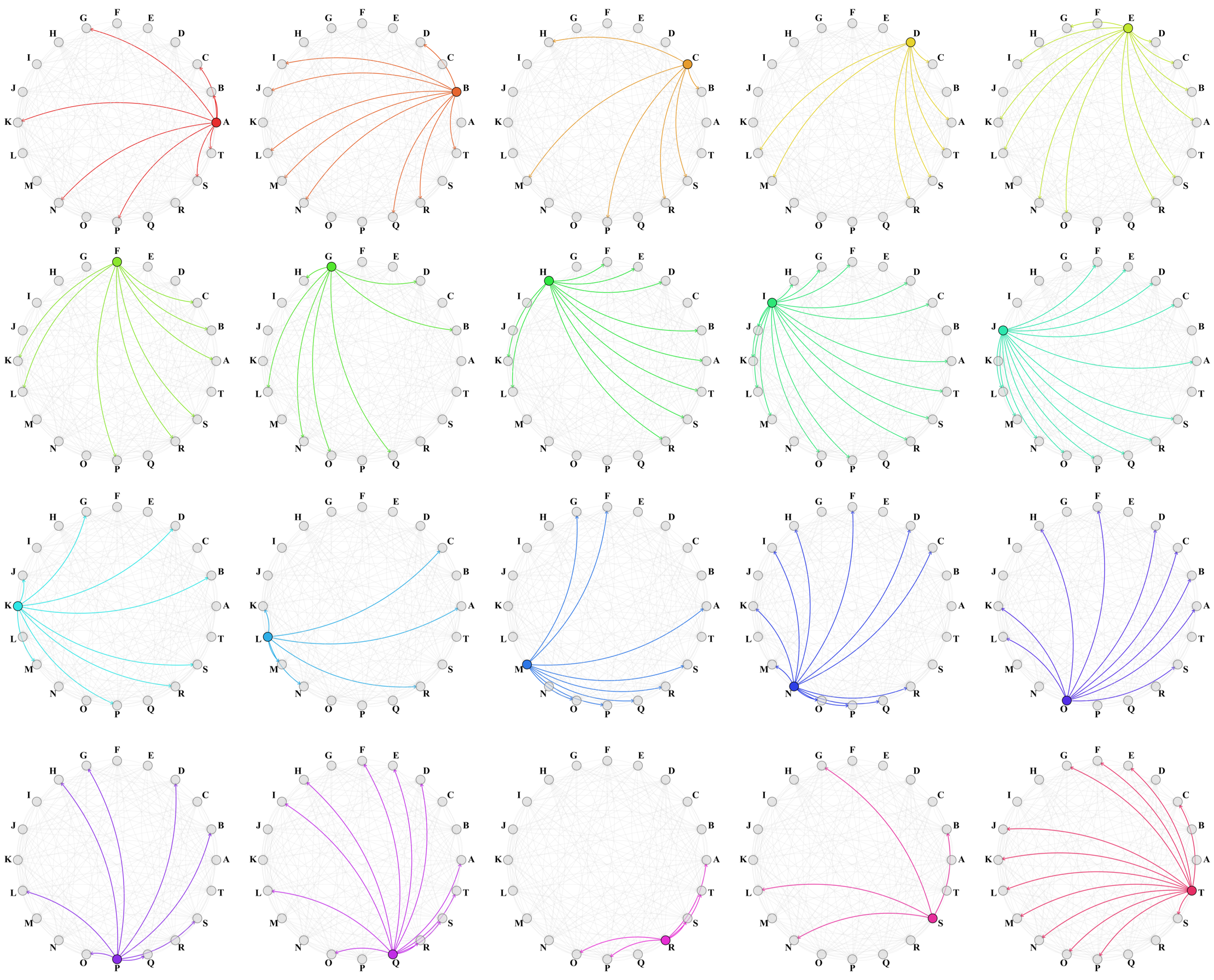}
    
    \caption{PLGs for Tic-Tac-Toe. The labels and colors of the nodes are consistent with Fig.~\ref{fig:plg}.}
    \label{plg:tictactoe}
    \vspace{-0.1cm}
\end{figure*}

\begin{figure*}[!h]
    \centering
    \includegraphics[width =0.91\linewidth]{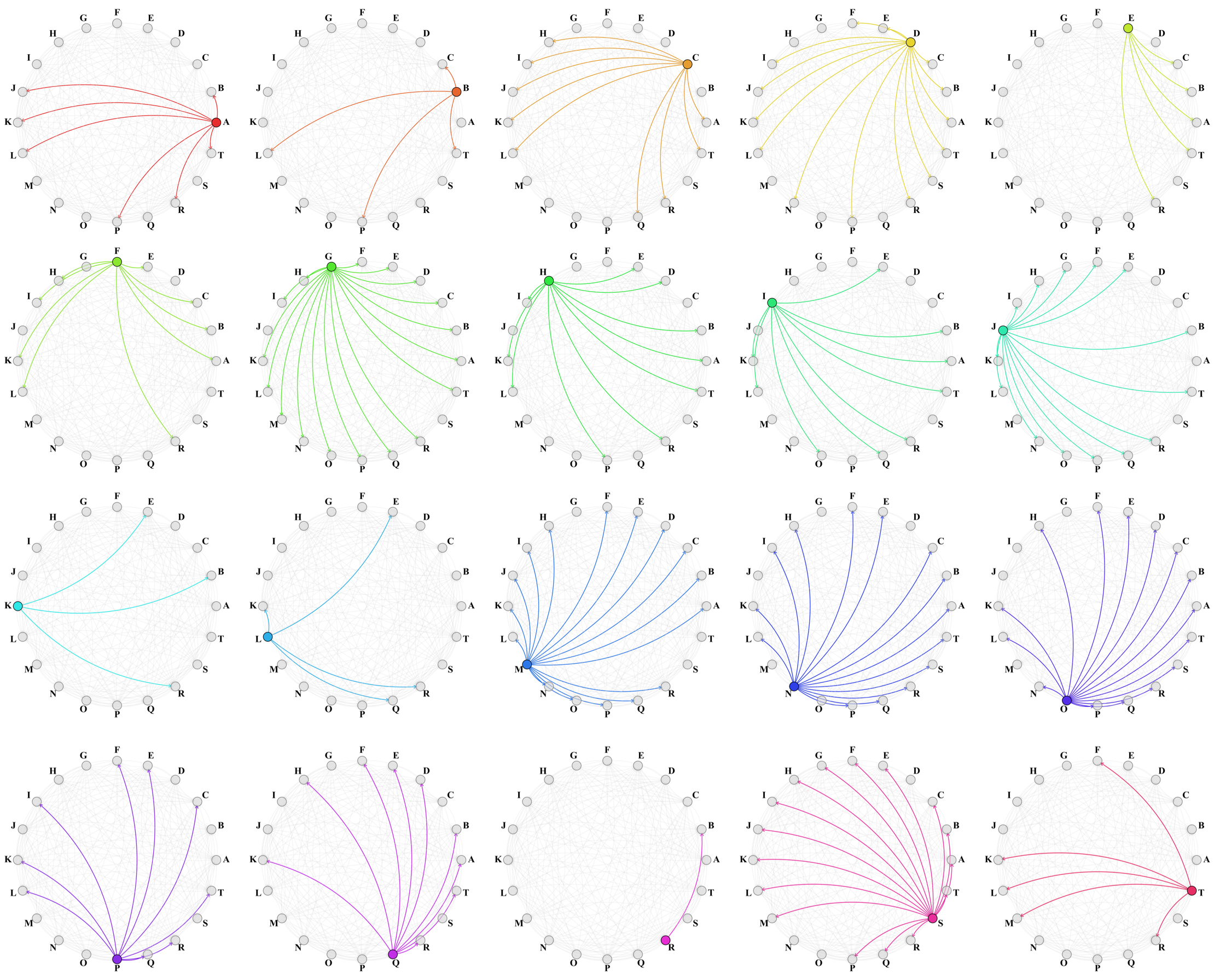}
    
    \caption{PLGs for Gomoku. The labels and colors of the nodes are consistent with Fig.~\ref{fig:plg}.}
    \label{plg:gomoku}
    \vspace{-0.1cm}
\end{figure*}

\begin{figure*}[!h]

    \centering
    \includegraphics[width =0.91\linewidth]{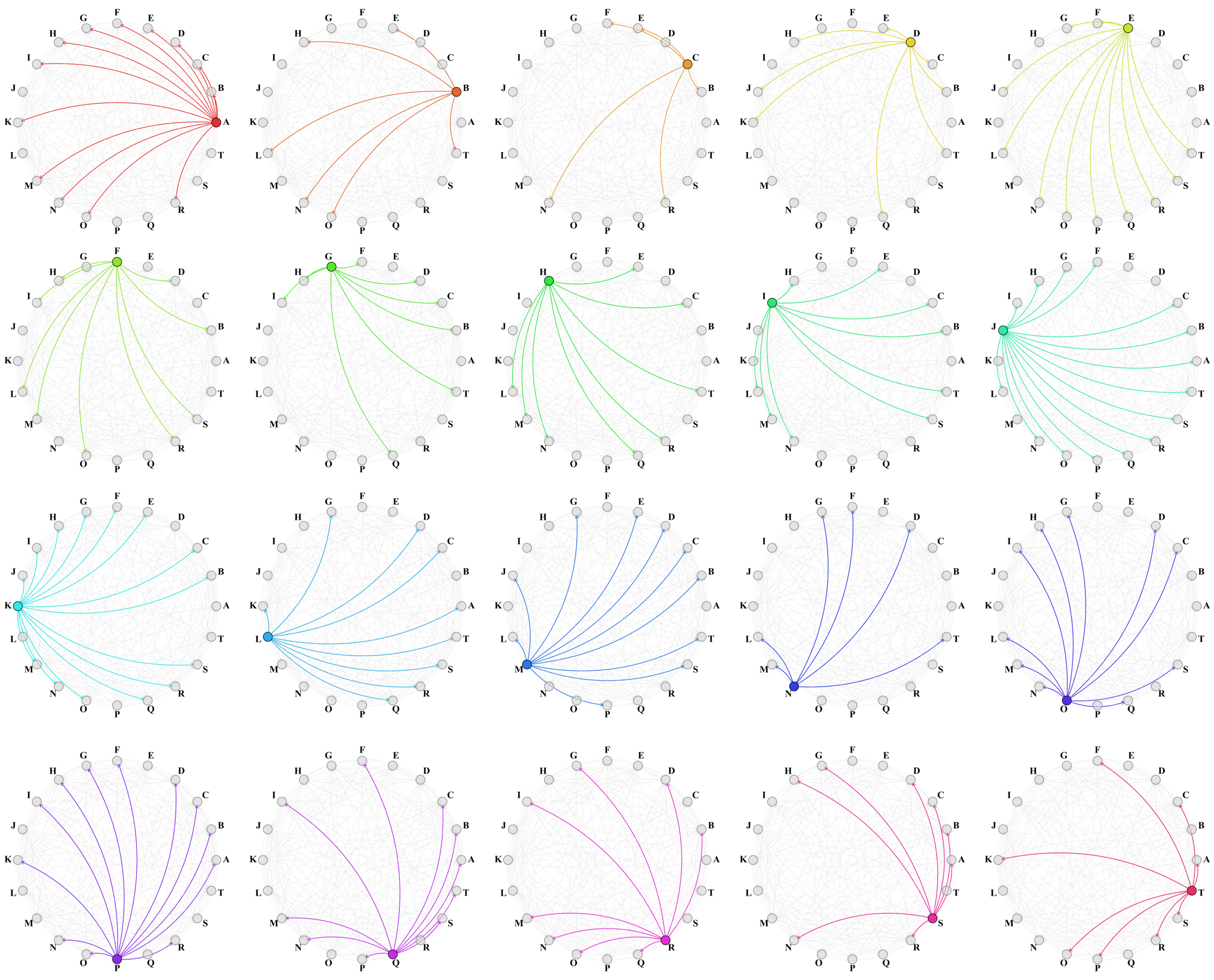}
    
    \caption{PLGs for Reversi. The labels and colors of the nodes are consistent with Fig.~\ref{fig:plg}.}
    \label{plg:reversi}
    \vspace{-0.1cm}
\end{figure*}

\begin{figure*}[!h]
    \centering
    \includegraphics[width =0.91\linewidth]{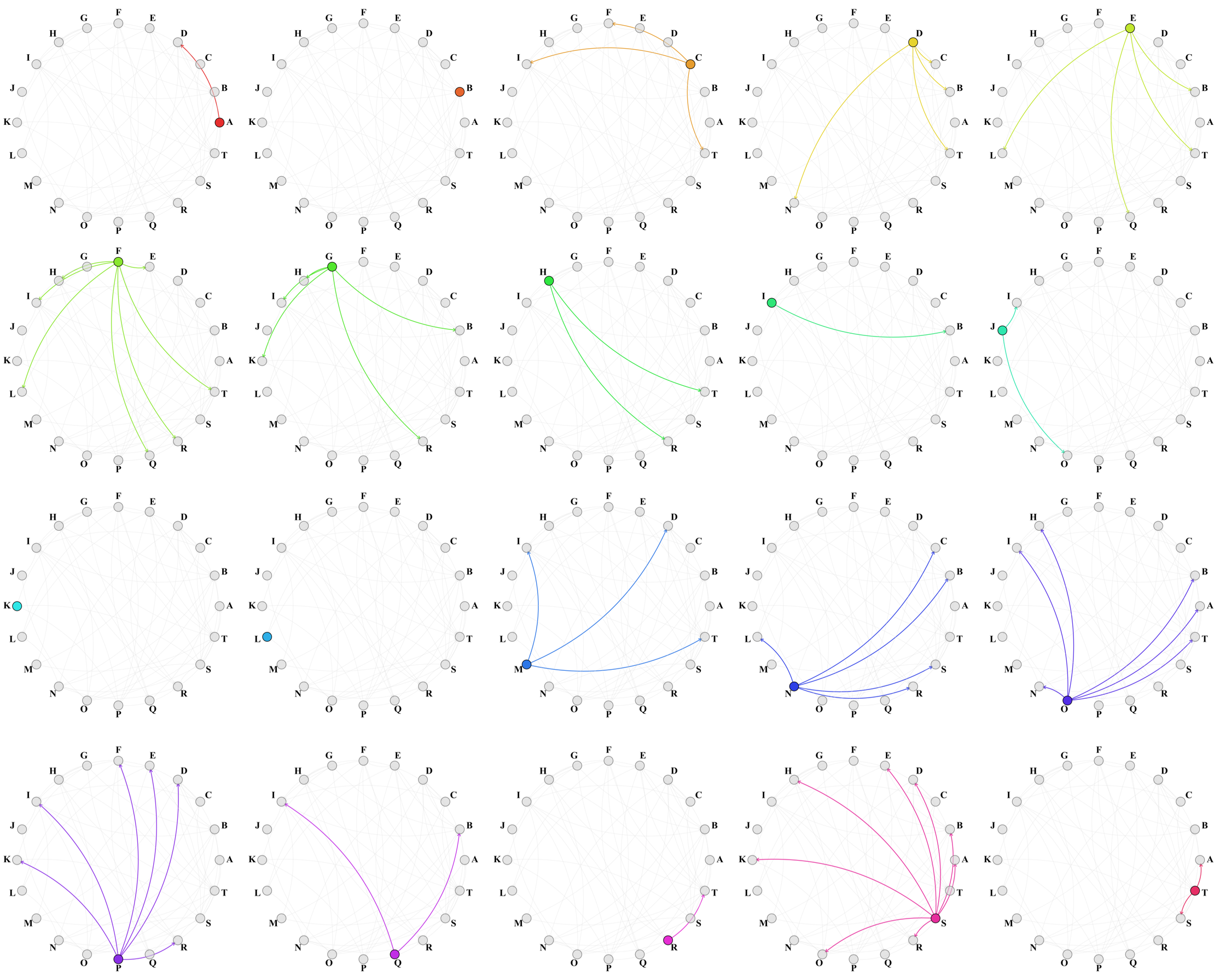}
    
    \caption{PLGs for Chess.The labels and colors of the nodes are consistent with Fig.~\ref{fig:plg}.}
    \label{plg:chess}
    \vspace{-0.1cm}
\end{figure*}

\begin{figure*}[!h]
    \centering
    \includegraphics[width =0.91\linewidth]{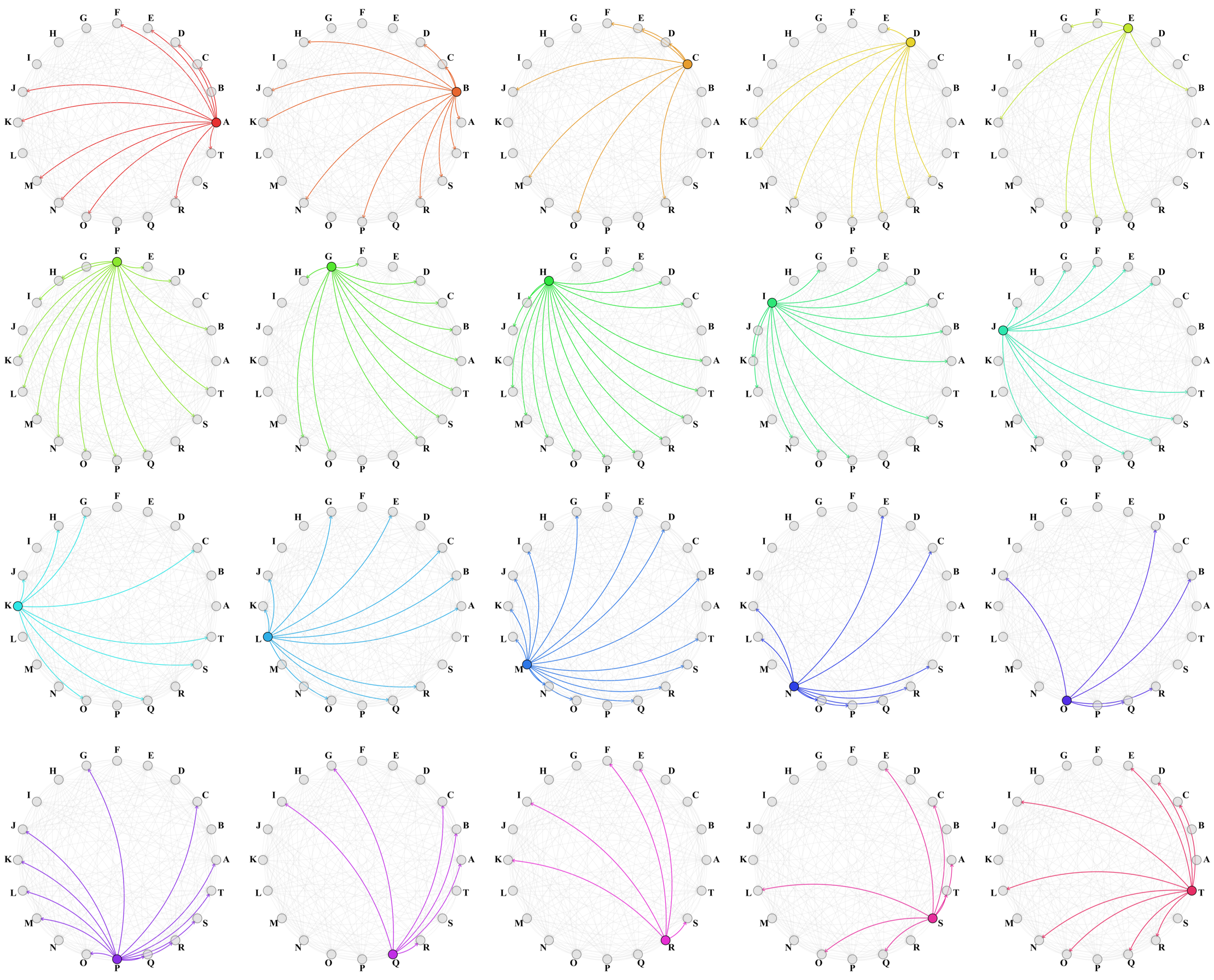}
    
    \caption{PLGs for Free-Style. The labels and colors of the nodes are consistent with Fig.~\ref{fig:plg}.}
    \label{plg:freestyle}
    \vspace{-0.1cm}
\end{figure*}

While Fig.~\ref{fig:plg} presents the Performance Loop Graphs (PLGs) corresponding to the top-performing player in each game category to facilitate intuitive understanding, the PLGs of other participating LLMs are not ignored. To provide a more comprehensive overview of our experimental findings, we now present the PLGs for all players across all game types, as illustrated in Figs.~\ref{plg:tictactoe} to \ref{plg:freestyle}. Several key observations can be derived from these figures: 1) PLGs offer a more visually intuitive representation of the win–loss dynamics among LLMs than the aggregate win/loss/draw statistics reported in Table~\ref{tab:performance}; 2) The PLGs associated with the same model differ substantially across game types, reinforcing the observation that LLMs exhibit inconsistent performance across different strategic environments; 3) The PLG visualizations reveal the presence of cyclical win–loss relationships among certain LLMs, thus forming closed loops. This phenomenon introduces a novel lens through which to evaluate the stability and relative significance of LLMs’ competitive capabilities.


\section{Temporal Analysis of Player Sentiment}
\label{sec:demo}
\begin{figure*}[!h]
    \vspace{1cm}
    \centering
    \includegraphics[width =0.98\linewidth]{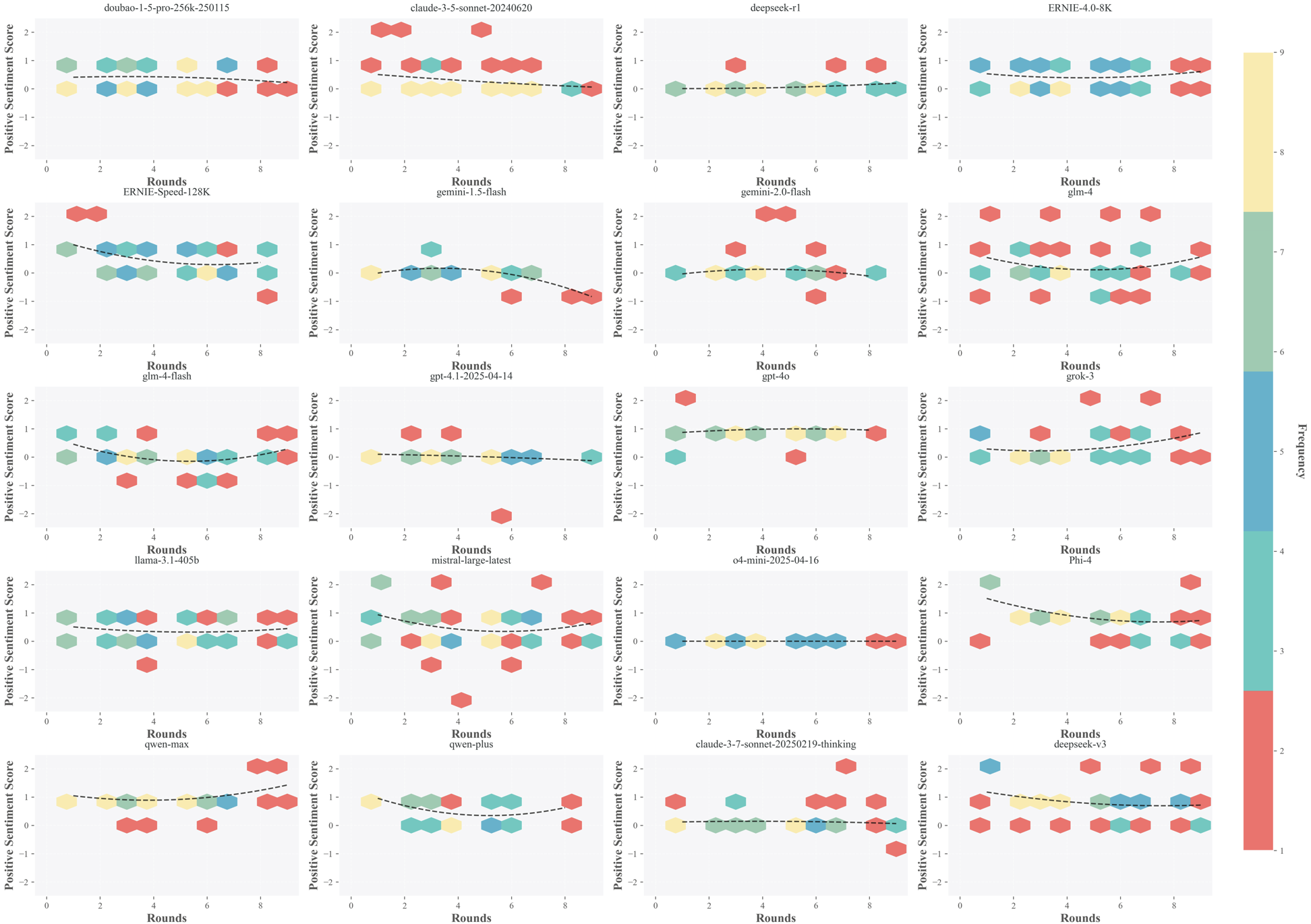}
    
    \caption{Emotional time-series heatmap for Tic-Tac-Toe.}
    \label{emo:tictactoe}
    \vspace{-0cm}
\end{figure*}

\begin{figure*}[!h]
    \centering
    \includegraphics[width =0.98\linewidth]{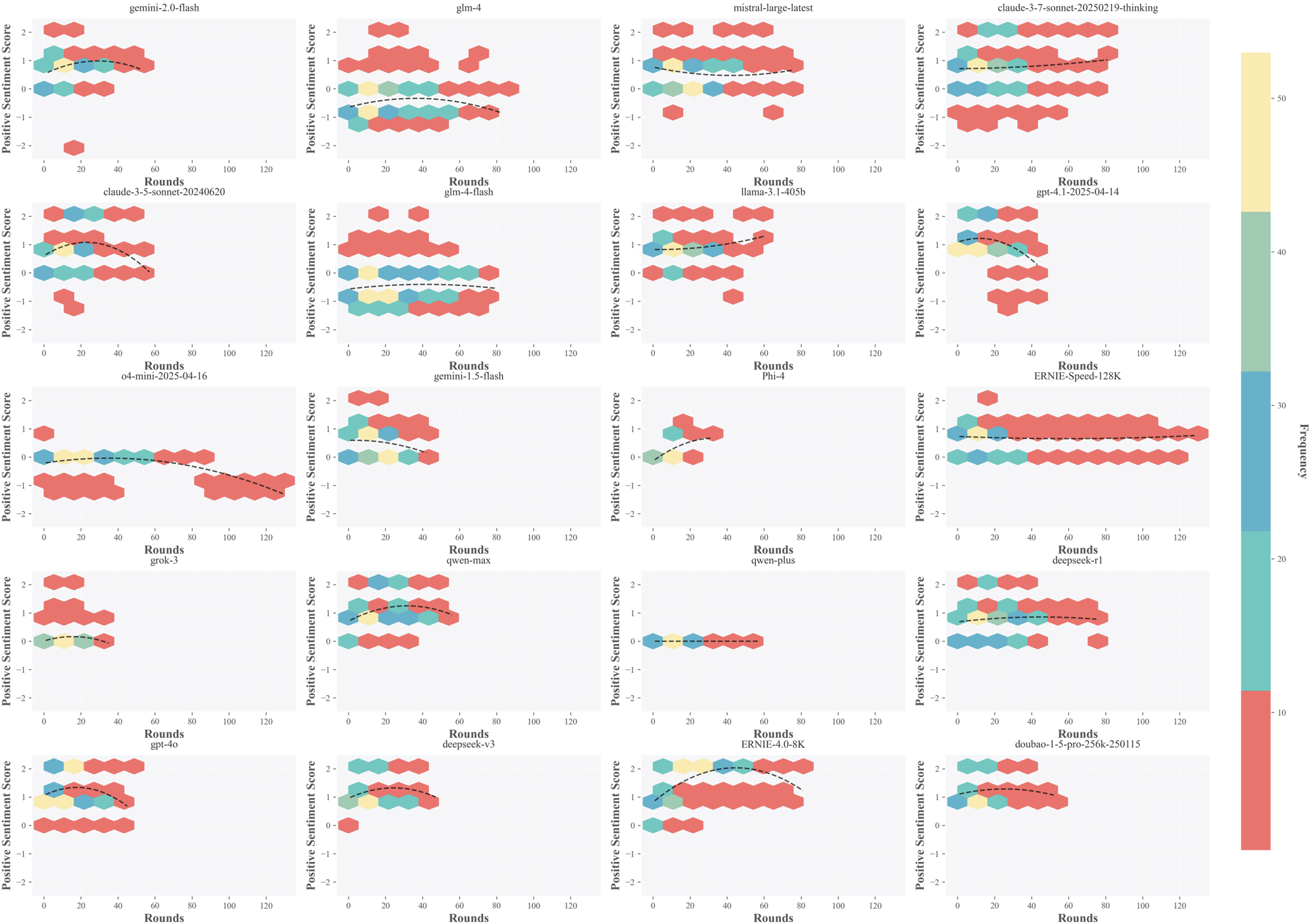}
    
    \caption{Emotional time-series heatmap for Gomoku.}
    \label{emo:gomoku}
    \vspace{0.8cm}
\end{figure*}

\begin{figure*}[!h]
    \centering
    \includegraphics[width =0.98\linewidth]{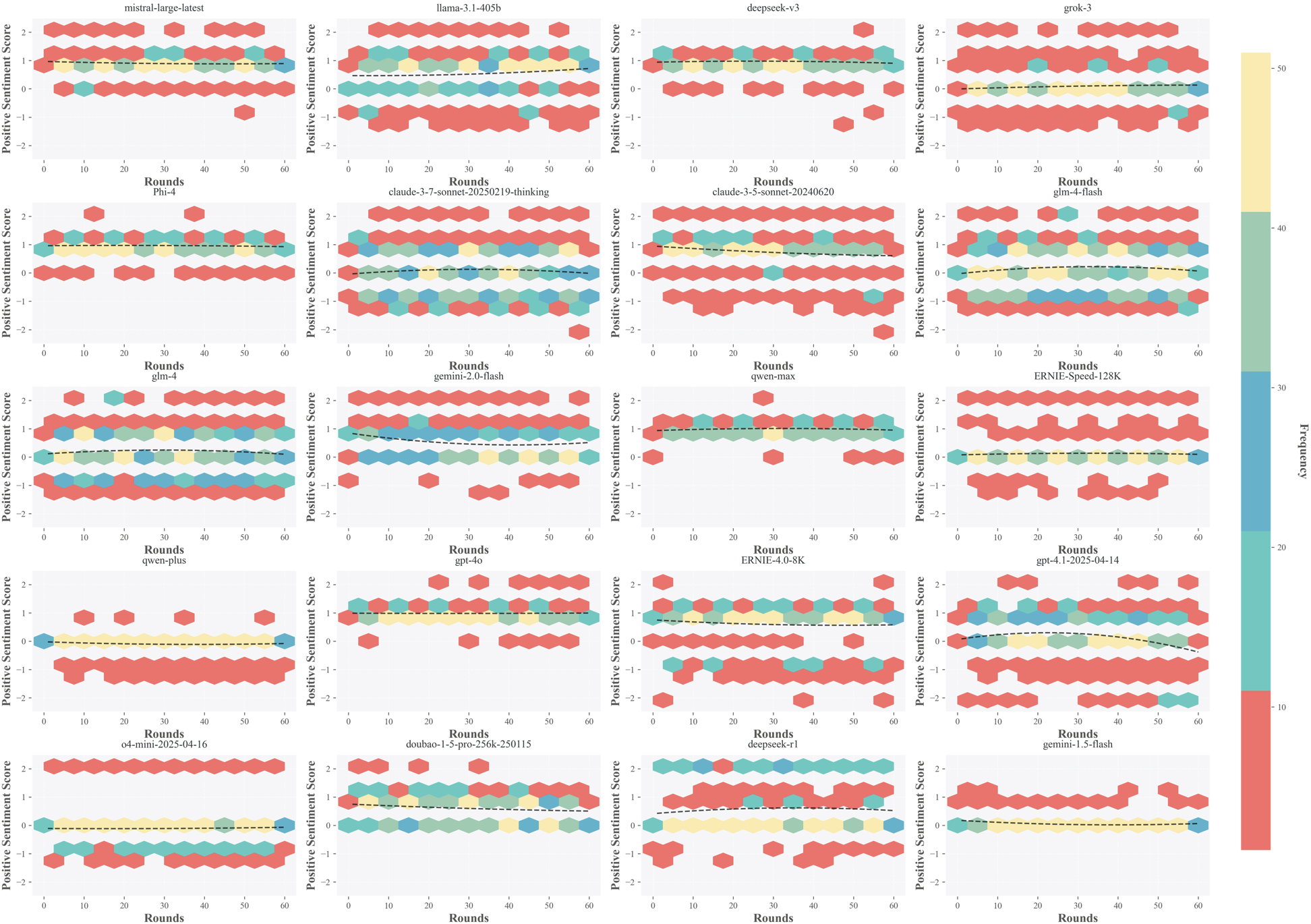}
    
    \caption{Emotional time-series heatmap for Reversi.}
    \label{emo:reversi}
    \vspace{-0.1cm}
\end{figure*}

\begin{figure*}[!h]
    \centering
    \includegraphics[width =0.98\linewidth]{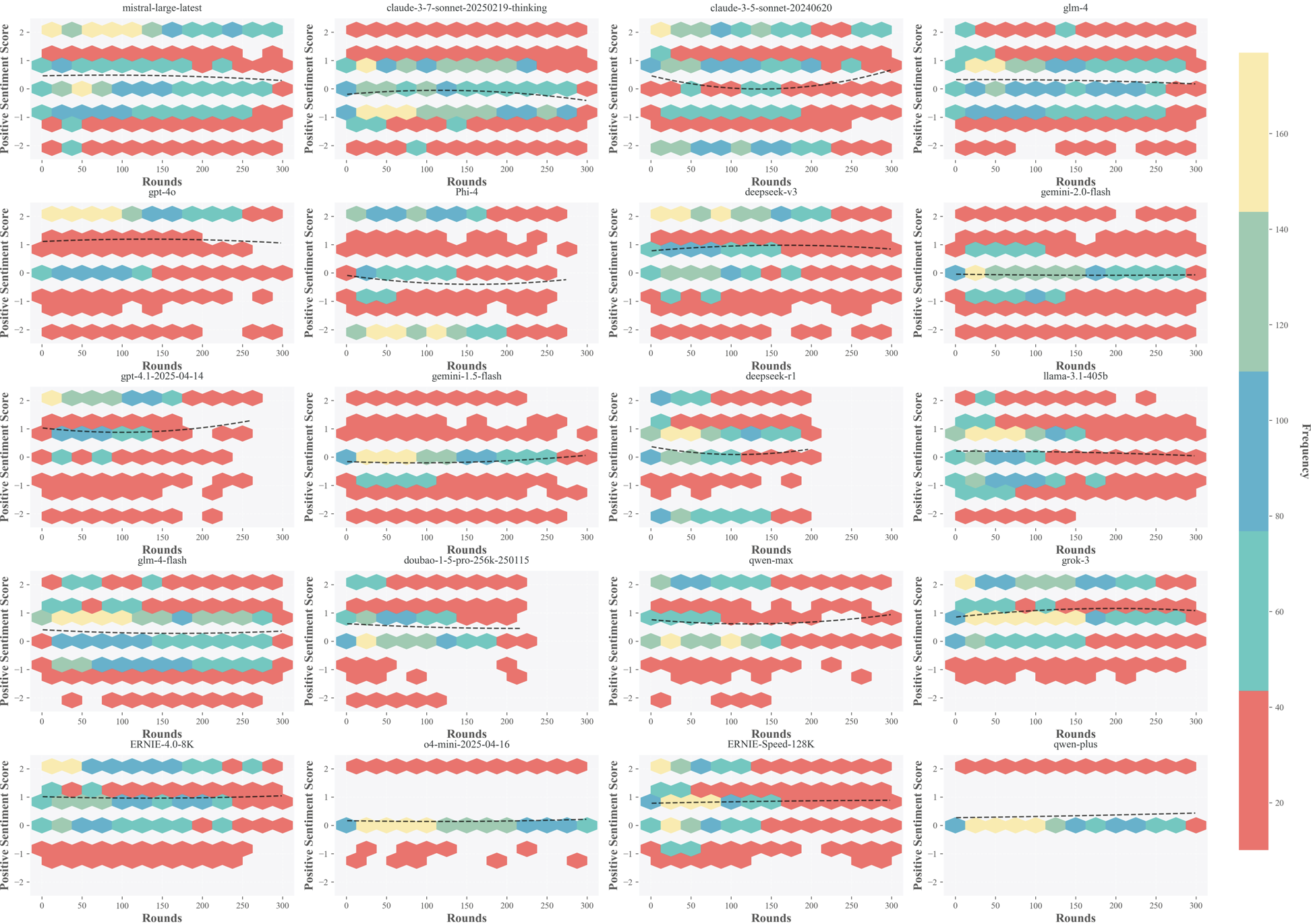}
    
    \caption{Emotional time-series heatmap for Chess.}
    \label{emo:chess}
    \vspace{0.8cm}
\end{figure*}

\begin{figure*}[!h]
    \centering
    \includegraphics[width =0.98\linewidth]{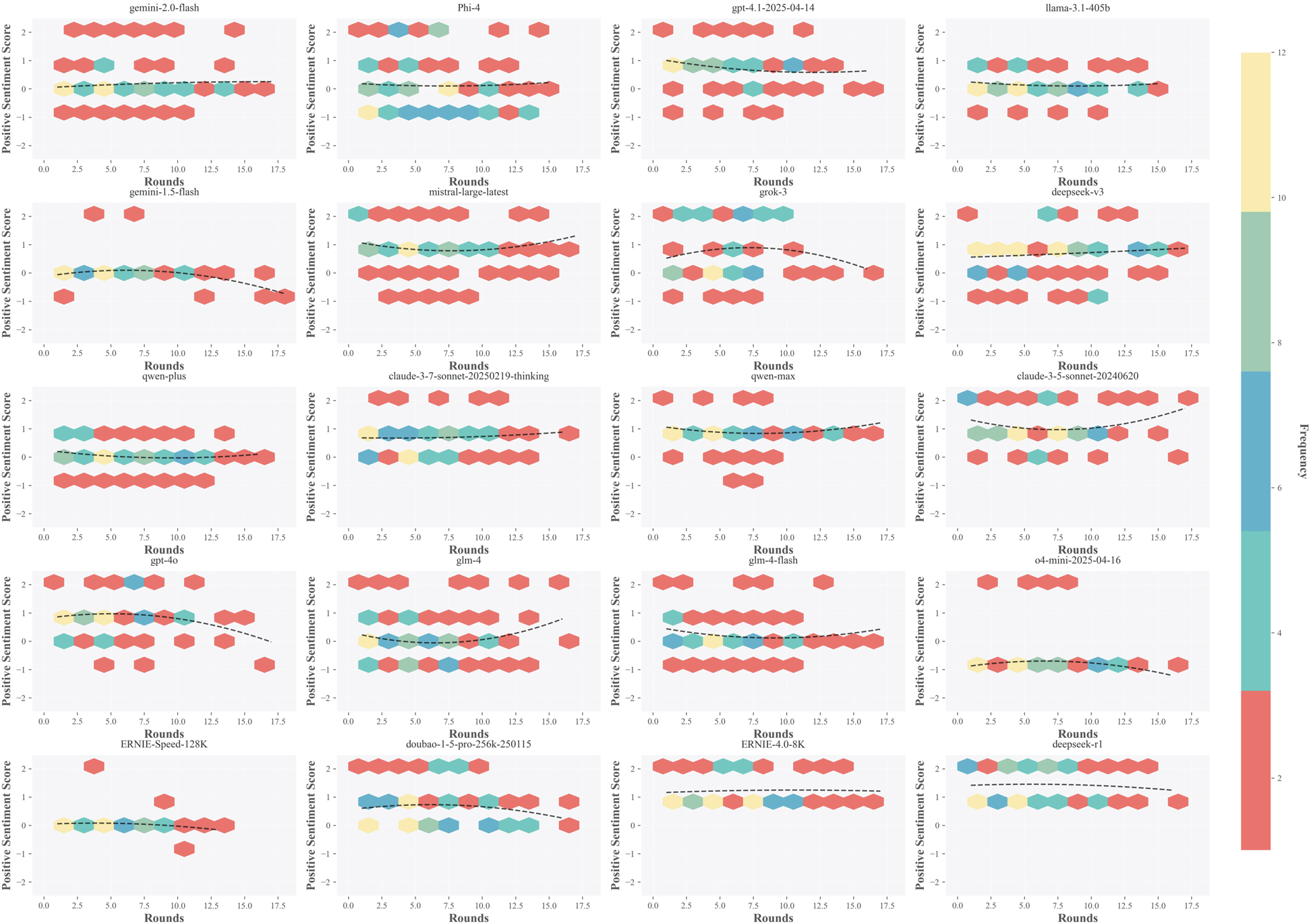}
    
    \caption{Emotional time-series heatmap for Free-Style.}
    \label{emo:freestyle}
    \vspace{-0.1cm}
\end{figure*}

Player emotion is inherently dynamic during gameplay, evolving in response to in-game events and strategic developments. As such, analyzing emotional characteristics solely through the static statistical methods outlined in Sec.~\ref{sec:emo} presents certain limitations. To further investigate the temporal dynamics of emotional responses in individual players, we introduce an emotional time-series heatmap visualization, shown in Figs~\ref{emo:tictactoe} to \ref{emo:freestyle}. In these visualizations, the horizontal axis represents the number of rounds played, the vertical axis corresponds to the Positive Sentiment Score, and the color intensity within each hexagonal bin indicates the frequency of specific emotional states over time. Beyond the findings reported in Sec.~\ref{sec:emo}, several additional insights emerge from the heatmaps: 1) Emotional trajectories differ markedly across game types. Notably, all LLMs tend to exhibit greater emotional volatility and more frequent negative sentiment when engaged in Chess, suggesting that cognitively demanding games exert a stronger influence on the emotional responses of LLMs compared to simpler games; 2) There is significant variation in the emotional dynamics across different models, even within the same game category. This heterogeneity may reflect underlying differences in the affective tendencies or simulated personality traits of the LLMs; 3) In addition to providing emotional data, the heatmaps also convey information about game duration. For instance, in Fig.~\ref{emo:gomoku}, both ERNIE-Speed and o4-mini extend their Gomoku matches beyond 120 rounds, while most other players reach a conclusion before 100 rounds. Similar observations across other game types offer further insight into each LLM's ability to manage gameplay and adapt strategy over time.

\end{document}